\documentclass[preprint,12pt]{elsarticle}

\usepackage{amssymb}
\usepackage{amsmath}
\usepackage{cases}
\usepackage{subfigure}
\usepackage{graphicx}
\usepackage{multirow}
\usepackage{mathrsfs}
\usepackage{enumitem}
\usepackage{bm}
\usepackage{amsthm}
\usepackage{mathtools}
\usepackage{booktabs} 
\usepackage{hyperref}
\usepackage{color}

\usepackage[ruled]{algorithm2e}
\newtheorem{theorem}{Theorem}

\newtheorem{definition}{Definition}

\usepackage{lineno}

\journal{Nuclear Physics B}

\begin{document}

\begin{frontmatter}



\title{ProLFA: Representative Prototype Selection for Local Feature Aggregation}


\author[mymainaddress,mysecondaddress]{Xingxing Zhang}
\ead{zhangxing@bjtu.edu.cn}

\author[mymainaddress,mysecondaddress]{Zhenfeng Zhu}
\ead{zhfzhu@bjtu.edu.cn}

\author[mymainaddress,mysecondaddress]{Yao Zhao\corref{mycorrespondingauthor}}
\cortext[mycorrespondingauthor]{Corresponding author}
\ead{yzhao@bjtu.edu.cn}


\address[mymainaddress]{Institute of Information Science, Beijing Jiaotong University, Beijing 100044, China}
\address[mysecondaddress]{Beijing Key Laboratory of Advanced Information Science and Network Technology}

\begin{abstract}
Given a set of hand-crafted local features, acquiring a global representation via aggregation is a promising technique to boost computational efficiency and improve task performance. Existing feature aggregation (FA) approaches, including Bag of Words and Fisher Vectors, usually fail to capture the desired information due to their pipeline mode. In this paper, we propose a generic formulation to provide a systematical solution (named \textbf{ProLFA}) to aggregate local descriptors. It is capable of producing compact yet interpretable representations by selecting representative prototypes from numerous descriptors, under relaxed exclusivity constraint. Meanwhile, to strengthen the discriminability of the aggregated representation, we rationally enforce the domain-invariant projection of bundled descriptors along a task-specific direction. Furthermore, ProLFA is also provided with a powerful generalization ability to deal flexibly with the semi-supervised and fully supervised scenarios in local feature aggregation. Experimental results on various descriptors and tasks demonstrate that the proposed ProLFA is considerably superior over currently available alternatives about feature aggregation.
\end{abstract}


\end{frontmatter}


\section{Introduction}

A typical visual task (e.g. classification or retrieval), which uses handcrafted features, consists of the following components: local feature extraction (e.g. SIFT~\cite{lowe2004distinctive} \textcolor{black}{and} DAISY~\cite{winder2009picking}), local feature aggregation (e.g. Bag of Words~\cite{csurka2004visual} \textcolor{black}{and} Fisher Vectors~\cite{perronnin2010improving}), and classification/retrieval/regression of the aggregated representations. This work focuses on the second component, \emph{i.e.,} the produce of compact yet \textcolor{black}{representative} representations from the local features. The aggregated results can not only reduce the computation memory of post-processing~\cite{jegou2012aggregating}, but also acquire the most valuable information of each sample. Furthermore, the performance of classification or retrieval task can be dramatically improved.

 
 \textcolor{black}{
 The problem of feature aggregation, generally referring to encoding and pooling of a series of local descriptors, has been well-studied in the literature~\cite{csurka2004visual,perronnin2010improving,huang2014feature}.
 Specifically, given a set of local descriptors, we first obtain a codebook by clustering all the descriptors, where the codebook is actually the set of cluster centers (also named codewords). Then, for each descriptor, we can find its most related codewords, and the statistics with respect to these codewords. Based on such information, each descriptor can be encoded as a new descriptor. Finally, we can obtain a global representation for the entire image by pooling the new descriptors belonging to that image.}
  For example, Bag of Words (BoW)~\cite{csurka2004visual} firstly quantizes every local descriptor according to a codebook that is commonly learned with K-Means~\cite{hartigan1979algorithm}. Then BoW represents each image as a histogram of codewords. The success of BoW aggregation prompted several extensions, including Fisher Vector (FV) coding~\cite{perronnin2010improving}, Super Vector (SV) coding~\cite{zhou2010image}, Locality-constrained Linear (LL) coding~\cite{wang2010locality}, Vector of Locally Aggregated Descriptors (VLAD)~\cite{jegou2010aggregating}, Vector of Locally Aggregated Tensors (VLAT)~\cite{picard2011improving}, and spatial fisher vectors~\cite{krapac2011modeling}. \textcolor{black}{By these approaches, a set of local features can be aggregated into a single vector.}

The success of these two-step methods, also called pipeline modes, naturally leads to the \underline{first} question---are there better encoding/pooling alternatives? \underline{Second}, the majority of existing methods completely rely on a visual codebook, which results in several attempts to improve the aggregation performance by improving the codebook. For example,~\cite{jurie2005creating} proposed a K-Means alternative that improves modelling of sparse regions of the local feature space.~\cite{jiu2012supervised} made direct use of the class labels in order to improve the BoW representation using a classifier. Recently, a spatially efficient yet accurate feature aggregation method~\cite{furuya2016aggregating} called Sum of Sparse Binary codes aggregation (SSB) is proposed, in which a set of sparse binary codes is aggregated by simple summing into a compact feature vector. Specifically, a family of local feature aggregation functions was defined in~\cite{katharopoulos2017learning} for any task that can be expressed as a differentiable cost function minimization problem. However, they still suffer from imperfect performance due to the negligence of the intrinsic structure among local descriptors. \underline{Third}, most of existing supervised aggregation methods are often particularly designed for a specific category recognition task, such as retrieval or classification~\cite{yang2008unifying,iosifidis2014discriminant,passalis2016entropy,passalis2018learning}, thus limiting their applications~\cite{wang2017multi}. We will address the mentioned three issues in following sections.

Additionally, to improve the interpretability of codebook, we consider the process of finding codebook as the task of \textbf{P}rototype \textbf{S}election (PS), which aims at finding exemplar samples from a feature collection. PS has been actively discussed in other fields~\cite{elhamifar2016dissimilarity}, such as video summarization and product recommendation, since it holds several advantages over data storage, compression, synthesis and cleansing. Besides helping to reduce the computational time and memory of algorithms, due to working on several prototypical samples, PS has further improved performances of numerous applications. Compared to dictionary learning methods such as K-Means~\cite{hartigan1979algorithm} and K-SVD~\cite{jiang2011learning,jiang2013label}, that learn centers/atoms in the input-space, PS methods~\cite{elhamifar2017subset} choose centers/atoms from the given samples, such as Kmedoids~\cite{kaufman1987clustering} and Affinity Propagation~\cite{frey2007clustering}.

\textcolor{black}{
In summary, the superiority of our prototype selection to conventional codebook learning lies in four key points. 
(i) Unlike those unsupervised codebook learning approaches (\emph{e.g.}, BoW~\cite{csurka2004visual} and VLAD~\cite{jegou2010aggregating}), the prototypes selected by our ProLFA are directly related to ultimate tasks (\emph{e.g.,} object recognition and image retrieval) by enforcing a task-specific projection. Thus, the aggregated features based on such prototypes are rich and discriminative enough to perform various tasks, yet compact to represent the entire image.
(ii) Our prototype selection can work out not only in fully supervised feature aggregation scenarios like those supervised codebook learning approaches (\emph{e.g.}, UniVCG~\cite{yang2008unifying} and DBoWs~\cite{iosifidis2014discriminant}), but also in semi-supervised scenarios. To the best of our knowledge, our work is the first attempt to obtain discriminative aggregated features in semi-supervised scenarios.
(iii) More importantly, instead of the popular used clustering strategy for codebook learning, we design an algorithmic feature aggregation formulation, where the diversity and representativeness of selected prototypes are explicitly formulated as an exclusivity constraint. Consequently, we can guarantee the quality of selected prototypes from the ultimate goal, which facilitates the interpretability of aggregated features on ultimate tasks.
(iv) In particular, our ProLFA can alleviate the influence of class unbalance on aggregated features, since the quality of selected prototypes can be enforced even in class unbalanced case.
}

\subsection{Contribution}
To the best of our knowledge, our work describes the \underline{first attempt} to jointly optimize the two key goals: FA interpretability and discrimination for final tasks.
In summary, the main contributions of this work are highlighted as follows.
\begin{itemize}[leftmargin=*]
   \item[i)] 
   We develop ProLFA: a \textbf{Pro}totype selection induced \textbf{L}ocal \textbf{F}eature \textbf{A}ggregation (ProLFA) model, to aggregate a set of  hand-crafted local features effectively for various tasks (\emph{e.g.,} image search and recognition).
   \item[ii)] The most representative prototypes are selected from numerous local descriptors under relaxed exclusivity constraint, thus facilitating the interpretability of aggregated representations.
   \item[iii)] By enforcing the domain-invariant projection of bundled descriptors along a task-specific direction, we can strengthen the discriminability of aggregated representations. Additionally, our model can deal flexibly with the semi-supervised and fully supervised scenarios in local feature aggregation.
   \item[iv)] A composite Block Coordinate Descent (cBCD) algorithm is customized to effectively seek the optimal solution of ProLFA. Experimental results have demonstrated our method works better than most of aggregation methods on a variety of features and tasks.
 \end{itemize}
  
\section{Related Work}
\textcolor{black}
{Depending on the order of statistics that connect codebook and local descriptors during encoding phase, feature aggregation approaches can be divided into two categories.
The first group of approaches mainly leverages first-order information, and usually encodes a local descriptor with weighted linear sum of related codewords prior to aggregation. 
Representative approaches include BoW~\cite{csurka2004visual}, DBoWs~\cite{iosifidis2014discriminant}, SC~\cite{ge2013sparse}, and LC-KSVD2~\cite{jiang2013label}. 
By contrast, the second group of approaches generally uses higher-order statistics (\emph{e.g.,} density, mean, and variances) that are computed from local descriptors and related codewords.
Although this kind of approach, such as FV~\cite{perronnin2010improving}, SV~\cite{zhou2010image}, VLAD~\cite{jegou2010aggregating}, and VLAT~\cite{picard2011improving}, takes more advantage, they still suffer from the influence of pipeline mode, where the codebook learning is independent from feature aggregation.
To address this issue, $\gamma$-democratic~\cite{lin2018second} exploited the relationship between democratic pooling and spectral normalization in the context of second-order features, and then proposed an aggregation approach in an end-to-end manner.
In addition, based on shallow aggregation approaches, several neural network based aggregation methods have also been proposed. FV+NN~\cite{perronnin2015fisher} proposed a hybrid architecture for image classification that took the advantages of FV~\cite{perronnin2010improving} and deep convolutional neural network (CNN) pipelines. It dramatically improved over previous FV systems without incurring the high complexity with respect to CNNs. 
Likewise, inspired by VLAD~\cite{jegou2010aggregating}, NetVLAD~\cite{arandjelovic2016netvlad} was proposed that is pluggable into any CNN architecture for weakly supervised tasks.
}
Our ProLFA can be considered as a combination of these two categories of FA approaches. This is due to: \textcolor{black}{(i)} Our ProLFA finds codebook by selecting prototypes from all local descriptors, where prototypes serve as dictionary. \textcolor{black}{(ii)} The density and mean of local descriptors are all involved in our ProLFA via the group term. 

According to whether auxiliary information about each sample is involved during codebook produce, existing FA models are often learned in an unsupervised or a supervised manner. 
Early representative approaches, such as BoW~\cite{csurka2004visual} and VLAD~\cite{jegou2010aggregating}, used unsupervised clustering algorithm to cluster the set of features and learn a dictionary. These unsupervised approaches achieved promising results and produced codebooks that were generic enough to be used for any \textcolor{black}{tasks}. However, learning a discriminative and task-oriented codebook is expected to perform significantly better. Therefore, some supervised aggregation approaches, such as $T_{1}\left ( \cdot  \right )$~\cite{katharopoulos2017learning} and EO-BoW~\cite{passalis2016entropy} were proposed.
By supervised dictionary learning, such approaches produce discriminative codebooks that are useful for the given classification or retrieval task. However, they cannot achieve the optimal performances simultaneously on all tasks. For instance, 
$T_{1}\left ( \cdot  \right )$~\cite{katharopoulos2017learning} could aggregate a highly discriminative representation for classification tasks, but it is not optimal for retrieval since it severely distorts the similarity between images in order to gain discriminability. 
It is worth noting that, our ProLFA is designed for any tasks (\emph{e.g.,} image classification, retrieval, annotation, or question answering), but is able to deal flexibly with the semi-supervised and fully supervised scenarios in local feature aggregation.

\section{The Proposed Model}

In this section, we firstly develop a ProLFA model to produce a global representation from a set of local descriptors, and then derive the algorithm to solve ProLFA.

\subsection{Model Formulation}

Suppose that we have $m$ samples $\left \{ \left (\bm X_{i},\bm y_{i} \right ):i=1,\ldots,m \right \}$ (\textcolor{black}{\emph{e.g.,}} images or texts), where $\bm X_{i}\in \mathbb{R}^{d\times N_{i}}$ is the set of $N_{i}$ local descriptors in $\mathbb{R}^{d}$ extracted from the $i^{th}$ sample, and $\bm y_{i} \in \mathbb{R}^{c}$ is the corresponding response vector of the $i^{th}$ sample. Meanwhile, we denote $\bm X = \left [ \bm x_{1},\ldots,\bm x_{N} \right ]\in \mathbb{R}^{d \times N}$ as the set of $\left \{ \bm X_{i} \right \}_{i=1}^{m}$, where $N = \sum_{i=1}^{m}N_{i}$. It is obvious that there exists much redundancy and irrelevance among these local descriptors. In order to improve the performance of final tasks (\textcolor{black}{\emph{e.g.,}} classification or retrieval), and meanwhile save the computational time, producing a global representation $\bar{\bm x}_{i}\in \mathbb{R}^{\bar{d}}$ for the $i^{th}$ sample is necessary. For this end, we propose an aggregation function $\Psi (\bar{\bm x}_{1},\ldots,\bar{\bm x}_{m})$, whose minimization over all possible aggregated representation set $\bar{\bm X}=\left \{\bar{\bm x}_{1},\ldots,\bar{\bm x}_{m} \right \}$, \emph{i.e.,}
\begin{align}\label{equ:main model}
\begin{split}
& \min \limits_{\left \{\bar{\bm x}_{1},\ldots,\bar{\bm x}_{m} \right \}} \Psi (\bar{\bm x}_{1},\ldots,\bar{\bm x}_{m})
\end{split}
\end{align}
needs to achieve two goals of (i) maximizing the discrimination of $\bar{\bm X}$; (ii) enhancing the interpretability of $\bar{\bm X}$.

\begin{figure}[!t]
    \centering
    \graphicspath{{pics/}}
    \includegraphics[width = 5.0in]{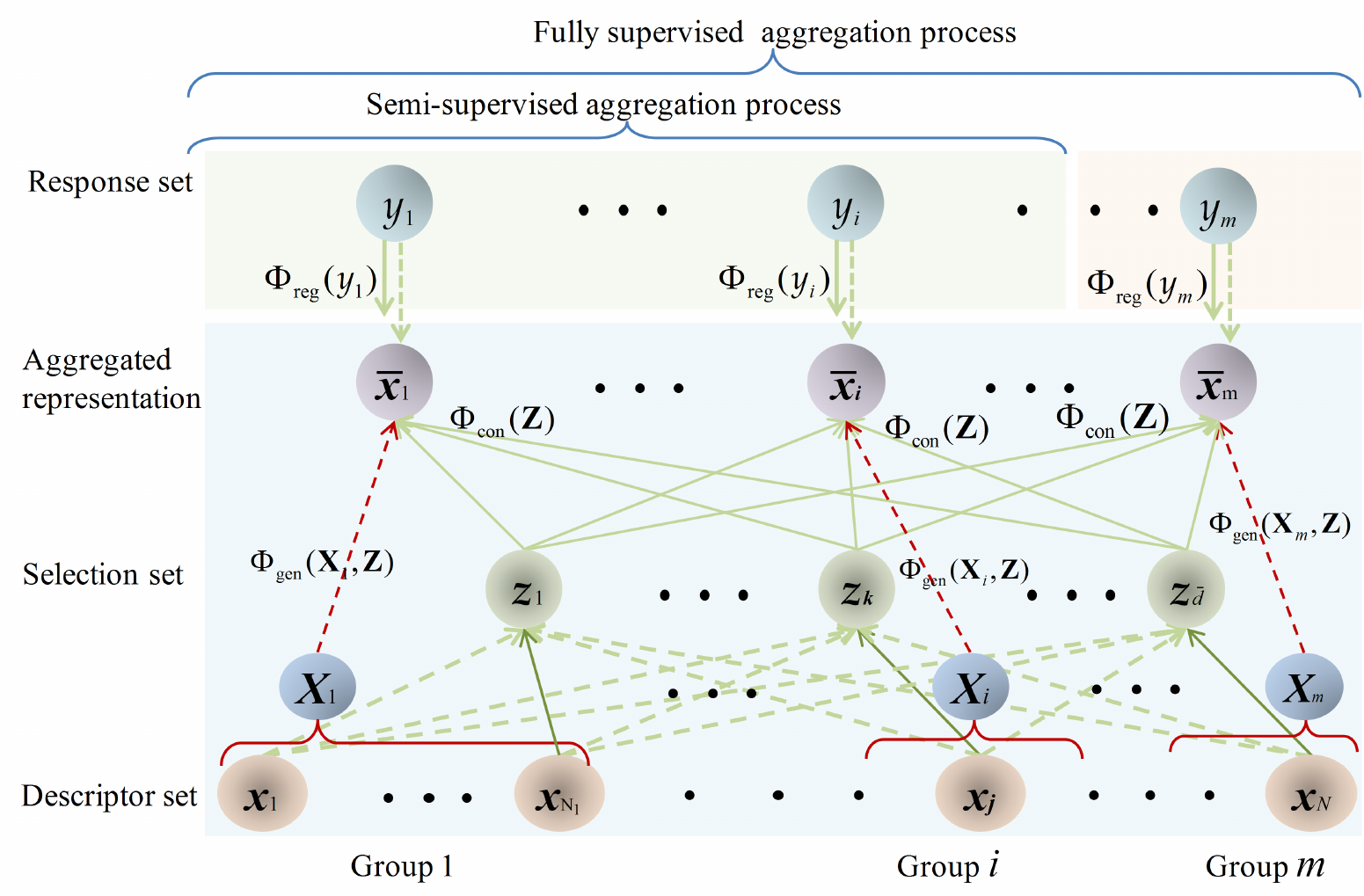}
    \caption{Illustration of the proposed ProLFA model. Given a collection of local descriptors, we can obtain the global representation for each sample by generation function $\Phi_{\rm{gen}}$. Specifically, representative prototypes are selected from numerous descriptors via selection matrix $\bm Z$, thus facilitating the interpretability of aggregated representations. 
    Furthermore, we impose the response vector set on the aggregated representations via regression function $\Phi_{\rm{reg}}$ to strengthen their discriminability.}
    \label{fig:overview}
    \vspace{-0.3cm}
    \end{figure}

As shown in Figure~\ref{fig:overview}, we consider a decomposition of the aggregation function $\Psi$ into two functions $\Phi_{\rm{reg}}$ and $\Phi_{\rm{gen}}$ with respect to the two aforementioned goals, as

\begin{align}\label{equ:sum}
\begin{split}
\Psi (\bar{\bm x}_{1},\ldots,\bar{\bm x}_{m}) := \sum_{i=1}^{m}\Phi_{\rm{reg}}( \underbrace{\Phi_{\rm{gen}}(\bm X_{i},\bm Z)}_{:=\bar{\bm x}_{i}}, \bm y_{i}; \bm W )+\Phi_{\rm{con}}(\bm W, \bm Z),
\end{split}
\end{align}
where $\bar{\bm x}_{i}=\Phi_{\rm{gen}}(\bm X_{i},\bm Z)$, $\bm Z$ is a prototype selection matrix, and $\Phi_{\rm{gen}}$ denotes the global representation \emph {generation function} that aims to produce interpretable representations by selecting the most prototypical local descriptors. $\bm W $ is a projection matrix from the feature space to the semantic space, and $\Phi_{\rm{reg}}$ denotes the \emph {regression function} that favors producing discriminative representations from $\bm X$ by maximally minimizing the regression error. $\Phi_{\rm{con}}$ represents the constraints imposed on $\bm Z$ and $\bm W $. Next we study each function \textcolor{black}{in~(\ref{equ:sum})}.

    \begin{figure*}[!t]
    \centering
    \graphicspath{{pics/}}
    \includegraphics[width = 5.4in]{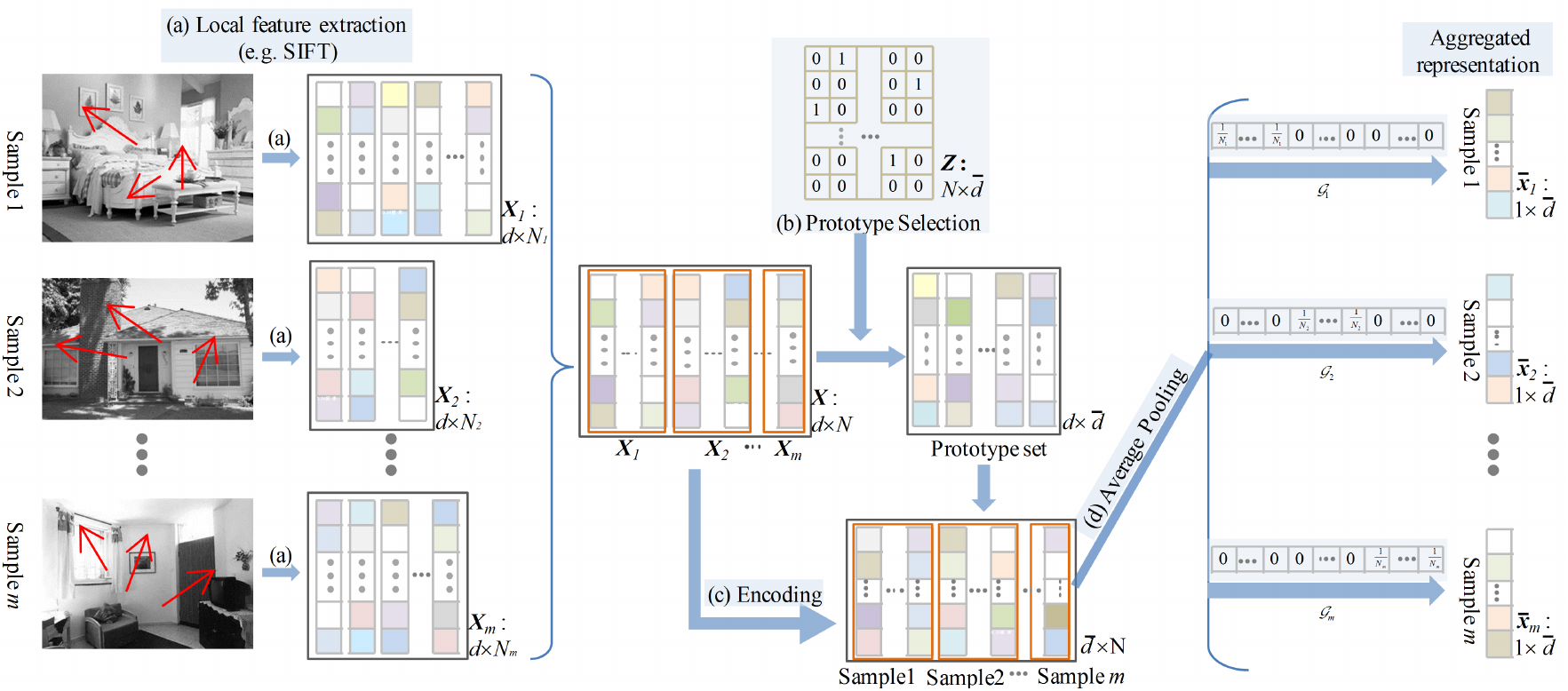}
    \caption{Framework of local feature aggregation by the function $\Phi_{\rm{gen}}$.}\label{fig:overFramework}
    \vspace{-0.3cm}
    \end{figure*}
 
\emph{\textbf{Generation Function.}} 
\textcolor{black}{
Instead of conventional codebook learning by clustering, we select the most prototypical descriptors to serve as codebook, where the diversity and representativeness of prototypes are guaranteed from the ultimate goal (\emph{e.g.,} object recognition and image retrieval). This can provide a clear physical meaning (\emph{i.e.,} interpretability) for aggregated features. For example, we can directly find their most related descriptors as well as locations, and meanwhile, aggregated features are representative enough to represent original images to perform the task. For this end, 
}
we integrate the codebook learning and feature encoding into a unified framework, instead of performing separately as in most existing aggregation approaches. Concretely, the generation function $\Phi_{\rm{gen}}$ is cast into a three layers network that can be formulated as:
\begin{align}\label{equ:generation}
\begin{split}
& \bar{\bm x}_{i}=\Phi_{\rm{gen}}(\bm X_{i},\bm Z)=\underbrace{\mathcal{\bm G}_{i}\underbrace{\bm X^{\rm T} \underbrace{\bm X \bm Z}_{\textbf{ \tiny{Prototypes}}}}_{\textbf{ \tiny{Encoding}}}}_{\textbf{ \tiny{Pooling}}},
\end{split}
\end{align}
where $\bm Z=\left [ \bm z_{1},\ldots,\bm z_{\bar{d}} \right ]\in \mathbb{R}^{N \times \bar{d}}$ denotes a selection matrix, $\bar{d}\ll N$, $\bm Z\in \left \{ 0,1 \right \}^{N\times\bar{d}}$ and $\bm 1^{\rm T} \bm Z=\bm 1^{\rm T}$.\footnote{$\bm 1$ denotes a vector, of appropriate dimension, whose elements are all equal to one.} $\mathcal{\bm G}_{i} \in \mathbb{R}^{N}$ aims to exploit intrinsic structure of $\bm X$ (including density and mean) by bundling the local descriptors in the $i^{th}$ sample. Depends on the bundle strategy, the construction of $\mathcal{\bm G}_{i}$ involves soft and hard forms\footnote{Soft and hard forms imply each element in $\mathcal{\bm G}_{i}$ is in the range $\left [ 0,\frac{1}{N_{i}} \right ]$ and $\left \{ 0,\frac{1}{N_{i}} \right \}$, respectively.}. In this work, we consider the hard one, \emph{i.e.,} for any sample $\bm X_{i}$, and $j=1,\cdots,N$,
\begin{equation}    \left ( \mathcal{\bm G}_{i} \right )^{j}:=
\begin{cases}
    \frac{1}{N_{i}},   &  \text{if $\bm x_{j}\in \bm X_{i}$}, \\
    0 ,     &  \text{otherwise}.
\end{cases}
\end{equation} \label{equ:ConstG}

As shown in Figure~\ref{fig:overFramework}, instead of clustering the set of local descriptors as in previous works~\cite{csurka2004visual}, the $\bar{d}$ selected prototypes with the proterities of diversity and representativeness serve as codebook, on which a compact representation $\bar{\bm x}_{i}$ is produced via average pooling.

\emph{\textbf{Regression Function.}} 
\textcolor{black}{
We introduce a domain-invariant projection, namely visual feature self reconstruction. Specifically, after projecting an aggregated feature vector into a semantic embedding space, it should be able to be projected back in the reverse direction to reconstruct itself. Such a strategy, similar to that used in autoencoder, can improve the model generalization ability as demonstrated in other problems~\cite{ap2014autoencoder,baldi2012autoencoders}.
}
Assuming that the forward and reverse projections have the same importance for feature aggregation, our regression function is then written as:
\begin{align}\label{equ:regression}
\begin{split}
 \Phi_{\rm{reg}}(\bar{\bm x}_{i}, \bm y_{i} ; \bm W )&=\left \|\bar{\bm x}_{i} \bm W-\bm y_{i}\right \|_{2}^{2}+\left \|\bar{\bm x}_{i} -\bm y_{i}\bm W ^{\textbf{T}}\right \|_{2}^{2}, 
\end{split}
\end{align}
where $\bm W \in \mathbb{R}^{\bar{d}\times c}$ denotes a projection matrix, and
$\bar{\bm x}_{i}$ is the $i^{th}$ normalized compact representation.

Our motivation can be explained as follows: 
\textcolor{black}{(i) Adding the losses of the forward and reverse projections imposes a self-reconstruction constraint on our regression function, which can improve the model generalization ability.} In our feature aggregation problem, this improved generalization ability makes the prototype selection matrix more applicable to the test samples.
\textcolor{black}{(ii)} By regression instead of typical classifiers, our ProLFA can perform various tasks besides classification, with different definitions of response vector $\bm y$, such as \emph{image annotation}, \emph{question answering}, and \emph{image classification} when $\bm y$ represents \emph{captions}, \emph{answers}, and \emph{labels}, respectively.

\emph{\textbf{Constraint Function.}} 
The constraint function $\Phi_{\rm{con}}$ is enforced both on the selection matrix $\bm Z$ and projection matrix $\bm W$. First, to characterize the representativeness of selected prototypes, we formulate an orthogonality constraint $\bm Z^{\rm T}\bm Z=\bm I$, \emph{i.e.,} $\left \| \bm z_{i}\odot \bm z_{j} \right \|_{0}=0$ for $i,j\in\left \{1,\cdots,\bar{d}\right \}$ and $i\neq j$, which leads to a diversified selection\footnote{$\odot$ designates the Hadamard product.}. Second, to enhance the stability of solution and mitigate the scale issue, we formulate a $\ell_{F}^{2}$ regularizer for $\bm W$, \emph{i.e.,} $\left \| \bm W \right \|_{F}^{2}$.

Using all the functions defined above, we can rewrite the \textbf{Pro}totype selection based \textbf{L}ocal \textbf{F}eature \textbf{A}ggregation model (ProLFA) in~(\ref{equ:main model}) as
\begin{align}\label{equ:model}
\begin{split}
& \min\limits_{\bm Z, \bm W}  \sum_{i=1}^{m}\left \|\mathcal{\bm G}_{i}\bm X^{\rm T}\bm X \bm Z \bm W-\bm y_{i}\right \|_{2}^{2}+ \sum_{i=1}^{m}\left \|\mathcal{\bm G}_{i}\bm X^{\rm T}\bm X \bm Z -\bm y_{i}\bm W^{\text{T}}\right \|_{2}^{2} + \lambda\left \| \bm W \right \|_{F}^{2}\\
&  s.t.\quad \bm 1^{\rm T}\bm Z =\bm 1^{\rm T};\quad \bm Z\in\left \{ 0,1 \right \}^{N\times\bar{d}};\quad \bm Z^{\rm T}\bm Z=\bm I,
\end{split}
\end{align}
where the regularization parameter $\lambda>0$ sets the trade-off between the three terms in the objective function. 

Due to the non-convexity and discontinuity of~(\ref{equ:model}), we have the following relaxation
\begin{align}\label{equ:ModelRelax}
\begin{split}
& \min\limits_{\bm Z, \bm W}  \sum_{i=1}^{m}\left \|\mathcal{\bm G}_{i}\bm X^{\rm T}\bm X \bm Z \bm W-\bm y_{i}\right \|_{2}^{2}+ \sum_{i=1}^{m}\left \|\mathcal{\bm G}_{i}\bm X^{\rm T}\bm X \bm Z -\bm y_{i}\bm W^{\text{T}}\right \|_{2}^{2}\\  &\quad \quad +2\lambda_{1} \underbrace{\sum_{j=1}^{\bar{d}}\sum_{i=1,i\neq j}^{\bar{d}}\left \| \bm z_{i}\odot \bm z_{j} \right \|_{1}}_{\textbf{Relaxed~Exclusivity}} +\lambda_{2}\left \| \bm W \right \|_{F}^{2} \\
&  s.t.\quad \bm 1^{\rm T}\bm Z =\bm 1^{\rm T};\quad \bm Z\geq \bm 0,
\end{split}
\end{align}
where $\lambda_{1}$ and $\lambda_{2}$ are nonnegative regularization parameters. Compared~(\ref{equ:ModelRelax}) with~(\ref{equ:model}), we first relax $\bm Z\in \left \{ 0,1 \right \}^{N\times\bar{d}}$ with $\bm Z\in \left [ 0,1 \right ]^{N\times\bar{d}}$. Second, instead of directly employing the constraint $\bm Z^{\rm T}\bm Z=\bm I$, we adopt the relaxed exclusivity constraint from a practical point of view, which is derived as follows:
\begin{align}\label{equ:relaxed exclusivity}
\begin{split}
\bm Z^{\rm T}\bm Z=\bm I &\Rightarrow \min \sum_{j=1}^{\bar{d}}\sum_{i=1,i\neq j}^{\bar{d}}\left \| \bm z_{i}\odot \bm z_{j} \right \|_{0}\Rightarrow \min \sum_{j=1}^{\bar{d}} \sum_{i=1,i\neq j}^{\bar{d}}\left \| \bm z_{i}\odot \bm z_{j} \right \|_{1}.
\end{split}
\end{align}

Therefore, the task of local feature aggregation is converted into an optimization program with respect to $\bm Z$ and $\bm W$.

\subsection{Optimization Framework} \label{section:Opti}

In order to efficiently solve the proposed ProLFA model in~(\ref{equ:ModelRelax}), we further rewrite it as the following equivalent
\begin{align}\label{equ:ModelRelax2}
 &\min\limits_{\bm Z, \bm W}  \mathcal {J} \left ( \bm Z, \bm W \right )  = \sum_{i=1}^{m}\left \|\mathcal{\bm G}_{i}\bm X^{\rm T}\bm X \bm Z \bm W-\bm y_{i}\right \|_{2}^{2}  + \sum_{i=1}^{m}\left \|\mathcal{\bm G}_{i}\bm X^{\rm T}\bm X \bm Z -\bm y_{i}\bm W^{\text{T}}\right \|_{2}^{2}\nonumber\\
 & \quad \quad + \lambda _{1}\left (\left \| \bm Z \right \|_{1,2}^{2}-\left \| \bm Z \right \|_{F}^{2}  \right ) +\lambda_{2}\left \| \bm W \right \|_{F}^{2}    \\
&  s.t.\quad \bm 1^{\rm T}\bm Z =\bm 1^{\rm T};\quad \bm Z\geq \bm 0,\nonumber
\end{align}
where the relaxed exclusivity term in~(\ref{equ:ModelRelax}) is replaced with the trick in Definition~\ref{def1}.
\begin{definition} \label{def1}
\begin{align}\label{equ:L12norm}
\begin{split}
\left \| \bm Z \right \|_{1,2}^{2}&:=\sum_{i=1}^{N}\left ( \sum_{j=1}^{\bar{d}}\left | z_{ij} \right | \right )^{2} 
=\left \| \bm Z\right \|_{F}^{2}+2\sum_{j=1}^{\bar{d}}\sum_{i=1,i\neq j}^{\bar{d}}\left \| \bm z_{i}\odot \bm z_{j} \right \|_{1}.
\end{split}
\end{align}
\end{definition}

The objective function in~(\ref{equ:ModelRelax2}) includes three convex terms and two mixed terms. Although not jointly convex in $\left ( \bm Z,\bm W \right )$, it is convex with respect to each unknown when the other is fixed. This is why Block Coordinate Descent (BCD) on $\bm Z$ and $\bm W$ performs reasonably well~\cite{mairal2009supervised}, although not necessarily providing the global optimum. A composite BCD (cBCD) solver consists therefore of iterating between \emph{Updating $\bm Z$} by fixing $\bm W$, and \emph{Updating $\bm W$} by fixing $\bm Z$. Below are the solutions to two subproblems.

\emph{\textbf{$\bm Z$ subproblem:}} If $\bm W$ is fixed, the subproblem in~(\ref{equ:ModelRelax2}) with respect to $\bm Z$ is written as
\begin{align}\label{equ:ModelZ}
\begin{split}
& \min\limits_{\bm Z}  \sum_{i=1}^{m}\left \|\mathcal{\bm G}_{i}\bm X^{\rm T}\bm X \bm Z \bm W-\bm y_{i}\right \|_{2}^{2} + \sum_{i=1}^{m}\left \|\mathcal{\bm G}_{i}\bm X^{\rm T}\bm X \bm Z -\bm y_{i}\bm W^{\text{T}}\right \|_{2}^{2}\\
&\quad \quad +\lambda _{1}\left (\left \| \bm Z \right \|_{1,2}^{2}-\left \| \bm Z \right \|_{F}^{2}  \right )   \\
&  s.t.\quad \bm 1^{\rm T}\bm Z =\bm 1^{\rm T};\quad \bm Z\geq \bm 0.
\end{split}
\end{align}

Considering the separability of both objective and constraints in~(\ref{equ:ModelZ}), we employ the Alternating Direction Method of Multipliers (ADMM) framework to solve this subproblem. To do so, we introduce an auxiliary matrix $\bm C\in \mathbb{R}^{N \times \bar{d}}$ and consider the following equivalent optimization program
\begin{align}\label{equ:ModelZ_v2}
\begin{split}
& \min\limits_{\bm Z, \bm C} \sum_{i=1}^{m}\left \|\mathcal{\bm G}_{i}\bm X^{\rm T}\bm X \bm C \bm W-\bm y_{i}\right \|_{2}^{2} + \sum_{i=1}^{m}\left \|\mathcal{\bm G}_{i}\bm X^{\rm T}\bm X \bm C -\bm y_{i}\bm W^{\text{T}}\right \|_{2}^{2}\\
&\quad \quad +\lambda _{1}\left (\left \| \bm Z \right \|_{1,2}^{2}-\left \| \bm Z \right \|_{F}^{2}  \right )  \\
&  s.t.\quad \bm 1^{\rm T}\bm C =\bm 1^{\rm T};\quad \bm C\geq \bm 0; \quad \bm Z =\bm C.
\end{split}
\end{align}

Augmenting the last equality constraint of~(\ref{equ:ModelZ_v2}) to the objective function via the Lagrange multiplier matrix $\bm \Lambda \in\mathbb{R}^{N \times \bar{d}}$ \textcolor{black}{and a positive penalty scalar $\mu$}, we can write the Lagrangian function as
\begin{align}\label{equ:Lagrange}
\begin{split}
\mathcal{L}\left ( \bm Z,\bm C,\bm \Lambda \right ) =&\sum_{i=1}^{m}\left \|\mathcal{\bm G}_{i}\bm X^{\rm T}\bm X \bm C \bm W-\bm y_{i}\right \|_{2}^{2}
+ \sum_{i=1}^{m}\left \|\mathcal{\bm G}_{i}\bm X^{\rm T}\bm X \bm C -\bm y_{i}\bm W^{\text{T}}\right \|_{2}^{2}\\
&+\frac{\mu}{2}\left \|\bm Z-\bm C\right \|_{F}^{2}
 +\left \langle \bm \Lambda,\bm Z-\bm C \right \rangle +\lambda _{1}\left (\left \| \bm Z \right \|_{1,2}^{2}-\left \| \bm Z \right \|_{F}^{2}  \right ).
\end{split}
\end{align}
\begin{itemize}[leftmargin=*]
  \item[-] \textcolor{black}{Minimizing~(\ref{equ:Lagrange})} with respect to $\bm Z$ can be done using an effective iteratively re-weighted algorithm~\cite{kong2014exclusive}. Concretely, we have
      \begin{align}\label{equ:OptimizeZ}
      \begin{split}
      \bm Z^{\left (t+1\right )}=&{\rm{argmin}}_{\bm Z}~ \lambda _{1}\left (\left \| \bm Z \right \|_{1,2}^{2}-\left \| \bm Z \right \|_{F}^{2}  \right )\\
      &+\frac{\mu}{2}\left \|\bm Z-\bm C^{\left (t\right )}\right \|_{F}^{2}+\left \langle \bm \Lambda^{\left (t\right )},\bm Z-\bm C^{\left (t\right )} \right \rangle.
      \end{split}
      \end{align}

      As observed from~(\ref{equ:OptimizeZ}), it can be split into $N$ independent smaller optimization programs over the $N$ rows of $\bm Z$. For each row vector $\bm Z_{\cdot j}$, we resolve the following equivalent objective:
      \begin{align}\label{equ:OptimizeZi}
      \begin{split}
      \bm Z_{\cdot j}^{\left (t+1\right )}= &{\rm{argmin}}_{\bm Z_{\cdot j}}~ \lambda _{1}\bm Z_{\cdot j}\bm F\bm Z_{\cdot j}^{\rm T}+\frac{\mu}{2}\left \|\bm Z_{\cdot j}-\bm C_{\cdot j}^{\left (t\right )}\right \|_{2}^{2} \\
      &+\left \langle \bm \Lambda_{\cdot j}^{\left(t\right )},\bm Z_{\cdot j}-\bm C_{\cdot j}^{\left (t\right )} \right \rangle,
      \end{split}
      \end{align}
      where $\bm F\in \mathbb{R}^{\bar{d} \times \bar{d}}$ is a diagonal matrix and formed by
      \begin{align}\label{equ:F}
      \begin{split}
      &\bm F:={\rm Diag}\left ( \left [\frac{\left \| \bm Z_{\cdot j} \right \|_{1}}{\left | \bm Z_{\cdot j}(1) \right |+\epsilon }-1,\cdots, \frac{\left \| \bm Z_{\cdot j} \right \|_{1}}{\left | \bm Z_{\cdot j}(\bar{d}) \right |+\epsilon } -1\right ] \right ),
      \end{split}
      \end{align}
      where $\epsilon \rightarrow 0^{+}$ (in the experiments, we use $10^{-1}$) is introduced to avoid zero denominators. With $\bm F$ fixed, by equating the partial derivative of~(\ref{equ:OptimizeZi}) with respect to $\bm Z_{\cdot j}$ to zero, we obtain
      \begin{align}\label{equ:Zi}
      \begin{split}
      &\bm Z_{\cdot j}^{\left (s+1\right )}=\left ( \mu\bm C_{\cdot j}^{\left (s\right )}-\bm \Lambda_{\cdot j}^{\left (s\right )} \right )\left (\mu I+2\lambda_{1}\bm F^{\left (s\right )} \right )^{-1}.
      \end{split}
      \end{align}
      Then $\bm F_{\cdot j}^{\left (s+1\right )}$ is updated using $\bm Z_{\cdot j}^{\left (s+1\right )}$ as in~(\ref{equ:F}). In an iterative way, the optimal value $\bm Z_{\cdot j}^{\left (t+1\right )}$ is obtained.

  \item[-] \textcolor{black}{Minimizing~(\ref{equ:Lagrange})} with respect to $\bm C$ subject to the probability simplex constraints $\left \{ \bm 1^{\rm T}\bm C =\bm 1^{\rm T},\bm C\geq \bm 0\right \}$ can be solved as follows:
      \begin{align}\label{equ:C}
      \begin{split}
       \bm C^{\left (t+1\right )} &= {\rm{argmin}}_{\left \{ \bm 1^{\rm T}\bm C =\bm 1^{\rm T},\bm C\geq \bm 0\right \}}~\mathcal{L}\left ( \bm C \right ) \\
      & \approx \mathop {\rm{argmin}}\limits_{\left \{ \bm 1^{\rm T}\bm C =\bm 1^{\rm T},\bm C\geq \bm 0\right \}} \left \| \bm C-\left.\left ( \bm C-\frac{1}{L}\frac{\partial \mathcal{L}\left ( \bm C \right )}{\partial \bm C} \right )\right|_{\bm C = \bm C^{\left (t\right )}}  \right \|_{F}^{2},
      \end{split}
      \end{align}
      where $\mathcal{L}\left ( \bm C \right )= \sum_{i=1}^{m}(\left \|\mathcal{\bm G}_{i}\bm X^{\rm T}\bm X \bm C \bm W^{\left (k\right )}-\bm y_{i}\right \|_{2}^{2}+ \left \|\mathcal{\bm G}_{i}\bm X^{\rm T}\bm X \bm C -\bm y_{i}\bm W^{\left (k\right )^{\text{T}}}\right \|_{2}^{2})
      +\frac{\mu}{2}\left \|\bm Z^{\left (t+1\right )}-\bm C\right \|_{F}^{2}+\left \langle \bm \Lambda^{\left (t\right )},\bm Z^{\left (t+1\right )}-\bm C \right \rangle$. $L$ is an upper bound of the Lipschitz constant of $\frac{\partial \mathcal{L}\left ( \bm C \right )}{\partial \bm C}$. By splitting~(\ref{equ:C}) into $\bar{d}$ independent smaller programs over the $\bar{d}$ columns of $\bm C$, \textcolor{black}{the algorithm}\footnote{The details are presented in Section 3 of~\cite{duchi2008efficient}.} in~\cite{duchi2008efficient} can be employed to solve each subproblem with respect to each column vector $\bm C_{i\cdot}^{\left (t+1\right )}$.

  \item[-] The multiplier matrix is updated by:
      \begin{align}\label{equ:Lambda}
      \begin{split}
      & \bm \Lambda^{\left (t+1\right )} = \bm \Lambda^{\left (t\right )}+\mu \left ( \bm Z^{\left (t+1\right )}-\bm C^{\left (t+1\right )} \right )\textcolor{black}{.}
      \end{split}
      \end{align}
 \end{itemize}

 For clarity, the procedure of solving the subproblem in~(\ref{equ:ModelZ}) is outlined in Algorithm~\ref{alg:Framwork1}. Convergence is achieved when we have $\left \| \bm Z^{\left (t+1\right )}-\bm C^{\left (t+1\right )} \right \|_{\infty}\leq \epsilon$ and $\left \| \bm Z^{\left (t+1\right )}-\bm Z^{\left (t\right )} \right \|_{\infty}\leq \epsilon$.

  \begin{algorithm}[tb]
  \footnotesize
            \caption{$\bm Z$ and $\bm C$ Solver using ADMM}
            \label{alg:Framwork1}
            \KwIn{$\left \{ \left (\bm X_{i},\bm y_{i} \right )\right \}_{i=1}^{m}$, $\left \{ \mathcal{\bm G}_{i} \right \}_{i=1}^{m}$, $\bm W^{\left (k\right )}$, $\bm Z^{\left (k\right )}$, $\bm C^{\left (k\right )}$, $\mu$.}

            initialization $t\leftarrow 0$; $\bm\Lambda^{\left (t\right )}$; $\bm Z^{\left (t\right )} \leftarrow \bm Z^{\left (k\right )}$; $\bm C^{\left (t\right )} \leftarrow \bm C^{\left (k\right )}$\;
            \While{not converged}{
                \For{$j=0:N$}
                {
                   initialization $s\leftarrow 0$\;
                   \While{not converged}{
                    Update $\bm F^{\left (s+1\right )}$ via Eq.~(\ref{equ:F})\;
                    Update $\bm Z_{\cdot j}^{\left (s+1\right )}$ via Eq.~(\ref{equ:Zi})\;
                    $s\leftarrow s+1$\;
                   }
                }
                $\bm Z^{\left (t+1\right )} \leftarrow \bm Z^{\left (s\right )}$\;
                Update $\bm C^{\left (t+1\right )}$ via Eq.~(\ref{equ:C})\;
                Update $\bm\Lambda^{\left (t+1\right )}$ via Eq.~(\ref{equ:Lambda})\;
                $t\leftarrow t+1$\;
            }
            \KwOut{$\bm Z^{\left (k+1\right )}\leftarrow \bm Z^{\left (t\right )}$, $\bm C^{\left (k+1\right )}\leftarrow \bm C^{\left (t\right )}$. }
  \end{algorithm}

\emph{\textbf{$\bm W$ subproblem:}} Given $\bm C$, the subproblem in~(\ref{equ:ModelRelax2}) with respect to $\bm W$ is written as
      \begin{align}\label{equ:OptW}
      \begin{split}
      & \min\limits_{\bm W}~ \sum_{i=1}^{m}\left \|\mathcal{\bm G}_{i}\bm X^{\rm T}\bm X \bm C^{\left (k\right )} \bm W-\bm y_{i}\right \|_{2}^{2}+ \sum_{i=1}^{m}\left \|\mathcal{\bm G}_{i}\bm X^{\rm T}\bm X \bm C^{\left (k\right )} -\bm y_{i}\bm W^{\text{T}}\right \|_{2}^{2}\\
      &\quad \quad  +\lambda_{2}\left \| \bm W \right \|_{F}^{2}\textcolor{black}{.}
      \end{split}
      \end{align}
      By equating the partial derivative of~(\ref{equ:OptW}) with respect to $\bm W$ to zero, we obtain a linear equation as follows:
      \begin{align}\label{equ:W}
      \begin{split}
      \bm A^{\left (k\right )} \bm W^{\left (k+1\right )}+\bm W^{\left (k+1\right )}\bm B^{\left (k\right )}=\bm Q^{\left (k\right )}\textcolor{black}{,}
      \end{split}
      \end{align}
     where $\bm A^{\left (k\right )} = \sum_{i=1}^{m}\bar{\bm x}_{i}^{\left (k\right )^\text{T}}\bar{\bm x}_{i}^{\left (k\right )}+\lambda_{2} I$, $\bm B^{\left (k\right )}=\sum_{i=1}^{m}{\bm y}_{i}^{\text{T}}{\bm y}_{i}$, $\bm Q=2\sum_{i=1}^{m}\bar{\bm x}_{i}^{\left (k\right )^\text{T}}{\bm y}_{i}$, and $\bar{\bm x}_{i}^{\left (k\right )}=\mathcal{\bm G}_{i}\bm X^{\rm T}\bm X \bm C^{\left (k\right )}$. (\ref{equ:OptW}) is a Sylvester equation and it can be solved efficiently by the Bartels-Stewart algorithm~\cite{bartels1972solution}.

In summary, Algorithm~\ref{alg:Framwork2} shows the steps of the cBCD implementation of the ProLFA model in~(\ref{equ:ModelRelax2}). The algorithm should not be terminated until the change of objective value is smaller than a pre-defined threshold (\emph{e.g.,} $10^{-1}$). For a new sample $\bm X_{\rm{new}}\in \mathbb{R}^{d\times N_{\rm{new}}}$, we finally obtain its corresponding global representation $\bar{\bm x}_{\rm{new}}$ as follows
\begin{equation}  \label{equ:TestSample}
   \begin{split}
   \bar{\bm x}_{\rm{new}}= \Phi_{\rm{gen}}(\bm X_{\rm{new}},\bm Z^{\ast})=\mathcal{\bm G}_{\rm{new}}{\bm X_{\rm{new}}}^{\rm T}\bm X \bm Z^{\ast}\textcolor{black}{,}
   \end{split}
\end{equation}
where $\mathcal{\bm G}_{\rm{new}}$ is the weight information of $N_{\rm{new}}$ descriptors in this new sample.

        \begin{algorithm}[tb]
        \footnotesize
            \caption{ProLFA Implementation using cBCD}
            \label{alg:Framwork2}
            \KwIn{$\left \{ \left (\bm X_{i},\bm y_{i} \right )\right \}_{i=1}^{m}$, $\left \{ \mathcal{\bm G}_{i} \right \}_{i=1}^{m}$, $\lambda_{1}$, $\lambda_{2}$, $\bar{d}$.}

            initialization $k\leftarrow 0$; $\bm W^{\left (k\right )} $; $\bm Z^{\left (k\right )} = \bm C^{\left (k\right )}$\;
            \While{not converged}{
             Update $\bm Z^{\left (k+1\right )}$ and $\bm C^{\left (k+1\right )}$ via Algorithm~\ref{alg:Framwork1}\;
             Update $\bm W^{\left (k+1\right )}$ via Eq.~(\ref{equ:W})\;
             $k\leftarrow k+1$\;
            }
            \KwOut{$\bm Z^{\ast}\leftarrow \bm Z^{\left (k\right )}$, $\bm W^{\ast}\leftarrow \bm W^{\left (k\right )}$.}
       \end{algorithm}

\section{Extension to Semi-supervised Aggregation}

With the emergence of large-scale data, the available labeled (or annotated) samples are usually inadequate for some tasks. As shown in Figure~\ref{fig:overview}, only $n$ samples $\left \{ \bm X_{\mathbb{I}_{j}}:j=1,\ldots,n; \mathbb{I}_{j}\in \left \{ 1,\ldots, m \right \} \right \}$ among the whole dataset $\left \{ \bm X_{i}:i=1,\ldots,m \right \}$ are labeled (or annotated) with the corresponding response vectors $\left \{ \bm y_{\mathbb{I}_{j}} \right \}$, where $n \ll m$ generally. To deal flexibly with this semi-supervised scenario in local feature aggregation, our ProLFA model in~(\ref{equ:ModelRelax2}) is reformulated as follows
\begin{align}\label{equ:ModelSemi}
\begin{split}
& \min\limits_{\bm Z, \bm W}  \sum_{j=1}^{n}\left \|\mathcal{\bm G}_{\mathbb{I}_{j}}\bm {\mathbb X}^{\rm T}\bm X \bm Z \bm W-\bm y_{\mathbb{I}_{j}}\right \|_{2}^{2}+\sum_{j=1}^{n}\left \|\mathcal{\bm G}_{\mathbb{I}_{j}}\bm {\mathbb X}^{\rm T}\bm X \bm Z -\bm y_{\mathbb{I}_{j}} \bm W^{\text{T}} \right \|_{2}^{2}\\
&\quad \quad +\lambda _{1}\left (\left \| \bm Z \right \|_{1,2}^{2}-\left \| \bm Z \right \|_{F}^{2}  \right ) +\lambda_{2}\left \| \bm W \right \|_{F}^{2}   \\
&  s.t.\quad \bm 1^{\rm T}\bm Z =\bm 1^{\rm T};\quad \bm Z\geq \bm 0,
\end{split}
\end{align}
where $\bm {\mathbb X}$ and $\bm X$ denote the sets of $\left \{ \bm X_{\mathbb{I}_{j}} \right \}_{j=1}^{n}$ and $\left \{ \bm X_{i} \right \}_{i=1}^{m}$, respectively. $\mathcal{K} \left ( \bm {\mathbb X}, \bm X \right )=\bm {\mathbb X}^{\rm T}\bm X$ is essentially a linear kernel matrix. $\mathcal{\bm G}_{\mathbb{I}_{j}}$ is \textcolor{black}{weight} of each descriptor in the $\mathbb{I}_{j}$-th labeled (or annotated) sample $\bm X_{\mathbb{I}_{j}}$. By employing Algorithm~\ref{alg:Framwork2}, we can obtain the optimal solutions $\bm Z^{\ast} \in \mathbb{R}^{N \times \bar{d}}$ and $\bm W^{\ast} \in \mathbb{R}^{\bar{d} \times c}$.
Furthermore, the corresponding global representation set $\bar{\bm {\mathbb X}}_{\mathbb{U}}=\left \{\bar{\bm x}_{\mathbb{I}_{j}}\right \}_{j=n+1}^{m}$ for unlabeled or unannotated samples in embedding space is
   \begin{equation}  \label{equ:reptest}
   \begin{split}
   \bar{\bm x}_{\mathbb{I}_{j}}= \Phi_{\rm{gen}}(\bm X_{\mathbb{I}_{j}},\bm Z^{\ast})=\mathcal{\bm G}_{\mathbb{I}_{j}}{\bm {\mathbb X}_{\mathbb{U}}}^{\rm T}\bm X \bm Z^{\ast}\textcolor{black}{,}
   \end{split}
\end{equation}
where $\bm {\mathbb X}_{\mathbb{U}}$ is the set of unlabeled or unannotated samples $\left \{ \bm X_{\mathbb{I}_{j}} \right \}_{j=n+1}^{m}$.

\section{Discussions}
The convergence, complexity, and scalability are analysed in this section.
\emph{\textbf{Convergence Analysis.}} The convergence behavior of our proposed cBCD algorithm is summarized as Theorem~\ref{the1}.
\begin{theorem} \label{the1}
The sequence of $\left \{ \mathcal {J} \left ( \bm Z^{\left ( k \right )}, \bm W^{\left ( k \right )} \right ) \right \}$, {i.e.,} the energy of the objective in~(\ref{equ:ModelRelax2}), generated by the proposed cBCD optimizer (Algorithm~\ref{alg:Framwork2}) converges monotonically.
\end{theorem}
\emph{Proof}: In terms of energy, the optimization nature of BCD ensures that~\cite{guo2018low}:
\begin{equation}  \label{equ:conver}
   \begin{split}
  \mathcal {J} \left ( \bm Z^{\left ( k \right )}, \bm W^{\left ( k \right )} \right ) \geq \mathcal {J} \left ( \bm Z^{\left ( k+1 \right )}, \bm W^{\left ( k \right )} \right ) \geq \mathcal {J} \left ( \bm Z^{\left ( k+1 \right )}, \bm W^{\left ( k +1\right )} \right ).  \nonumber
   \end{split}
\end{equation}
In other words, the energy gradually decreases as the involved two steps iterate. Further, the whole objective function~(\ref{equ:ModelRelax2}) has a lower bound. Therefore, Algorithm~\ref{alg:Framwork2} is guaranteed to converge monotonically.

\textcolor{black}{\emph{\textbf{Complexity Analysis.}}} We consider using $P$ parallel processing resources to solve the proposed optimization problem in~(\ref{equ:ModelRelax2}). Thereby, updating each row of $\bm Z$ takes $\mathcal{O}(\alpha_{1} \bar{d})$ and $\mathcal{O}(\alpha_{1}{\bar{d}}^{2})$ for (\ref{equ:F}) and (\ref{equ:Zi}) respectively, where $\alpha_{1}$ is the (inner) iteration number in Algorithm~\ref{alg:Framwork1}. Specifically, due to the diagonalization of $\bm F$, the inverse operator in (\ref{equ:Zi}) only needs $\mathcal{O}(\bar{d})$. Updating $\bm C$ takes $\mathcal{O}\left ( N\left \lceil \bar{d}/P \right \rceil \right )$ for (\ref{equ:C}) using the randomized algorithm in~\cite{duchi2008efficient}. Therefore, the cost of Algorithm~\ref{alg:Framwork1} is $\mathcal{O}(\alpha_{2}(\alpha_{1}\bar{d}^{2}\left \lceil N/P \right \rceil+N \left \lceil \bar{d}/P \right \rceil))$, where $\alpha_{2}$ is the number of (outer) iterations required to converge. Given $\bm A$, $\bm B$ and $\bm Q$, solving $\bm W$ via (\ref{equ:W}) spends $\mathcal{O}(\bar{d}^{3}+c^{3})$. The computation of $\bar{\bm x}_{i}$, $\bm A$ and $\bm Q$ has a time complexity of $\mathcal{O}(\bar{d}N)$, $\mathcal{O}(\bar{d}^{2}m)$, and $\mathcal{O}(\bar{d}cm)$, respectively. Due to $N\gg\bar{d}>c$, Algorithm~\ref{alg:Framwork2} has the complexity of $\mathcal{O}(\alpha_{3}(\alpha_{2}(\alpha_{1}\bar{d}^{2}\left \lceil N/P \right \rceil+N \left \lceil \bar{d}/P \right \rceil)+\bar{d}(\bar{d}^{2}+\bar{d}m+N)))$, where $\alpha_{3}$ is the number of iterations required to converge. 


\emph{\textbf{Scalability Analysis.}} As observed from the complexity analysis, the complexity of our algorithm shows a roughly linearly increasing timing result, which is a general case for some typical works in feature aggregation (\emph{e.g.,} BoW). Thus, the proposed model has some limitations for directly dealing with a large set of local features. To alleviate this issue, we can resort to splitting the unlabeled (or unannotated) data $\bm {\mathbb X}_{\mathbb{U}}$ into multiple batches, on which prototypes can be selected recursively. Ideally, its feasibility and effectiveness by this way can be expected.

\section{Experimental Verification}

\subsection{Experimental Setup}

For a given image, we apply the Hessian-affine detector~\cite{jegou2014triangulation} to detect multiple features. For each detected local feature, we then compute two types of local descriptors, SIFT~\cite{lowe2004distinctive} and DAISY~\cite{winder2009picking}. The compared aggregation approaches include unsupervised ones (BoW~\cite{csurka2004visual}, VLAD~\cite{jegou2010aggregating}, FV~\cite{perronnin2010improving}, SC~\cite{ge2013sparse}, $\phi_{\Delta }+\psi_{d}+ \rm RN$~\cite{jegou2014triangulation}, and DM~\cite{furuya2015diffusion}). \textcolor{black}{Meanwhile,} five supervised aggregation approaches, $T_{1}\left ( \cdot  \right )$~\cite{katharopoulos2017learning}, UniVCG~\cite{yang2008unifying}, LC-KSVD2~\cite{jiang2013label}, DBoWs~\cite{iosifidis2014discriminant}, and EO-BoW~\cite{passalis2016entropy}, are also employed for comparison. It is worth noting that the proposed model focuses on the semi-supervised and supervised feature aggregation. However, most of existing typical works rely on unsupervision. Thus, we additionally compared some unsupervised ones to verify the effectiveness of the introduced response information. Conducting such a comparison is also common in existing supervised aggregation works, \emph{e.g.,} UniVCG~\cite{yang2008unifying} and Eo-BoW~\cite{passalis2016entropy}. For clarity, we denote our semi-supervised model in~(\ref{equ:ModelSemi}) as Semi-ProLFA.
To avoid the influence of randomness, we average the results over 6 times of execution with different training set selections. Additionally, we have tried to yield better performances of \textcolor{black}{all compared approaches} by tuning the related parameters.

\subsection{Synthetic Dataset}

\begin{figure}[!t]
  \setlength{\abovecaptionskip}{0pt}
  \centering
  \graphicspath{{pics/}}
  {
  \label{fig:kmeans}
  \includegraphics[width = 2.7in]{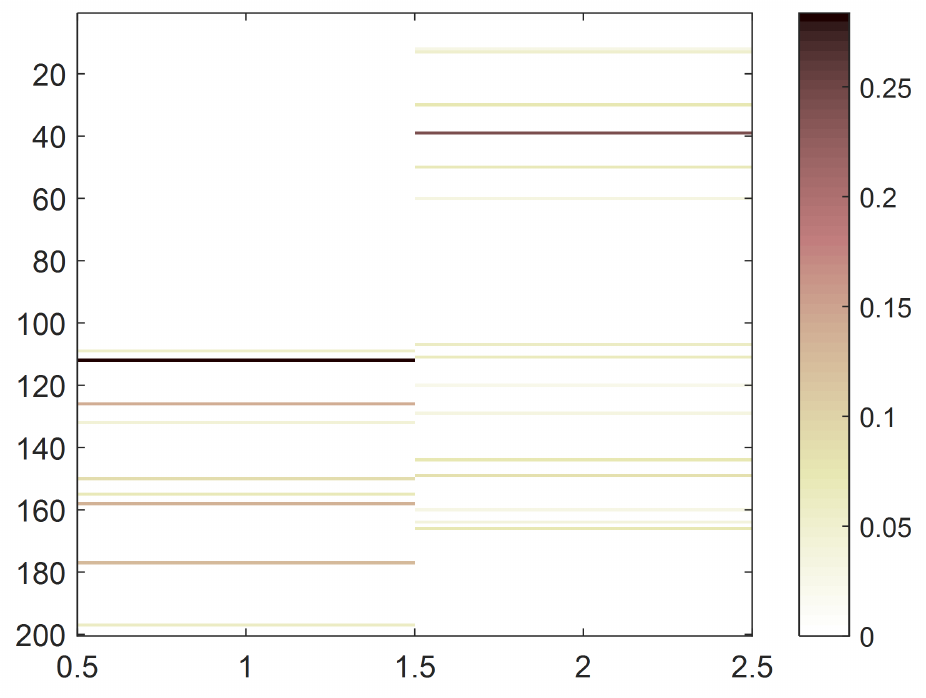}}
  \hspace{0.05cm}
  {
  \label{fig:ProLFA}
  \includegraphics[width = 2.5in]{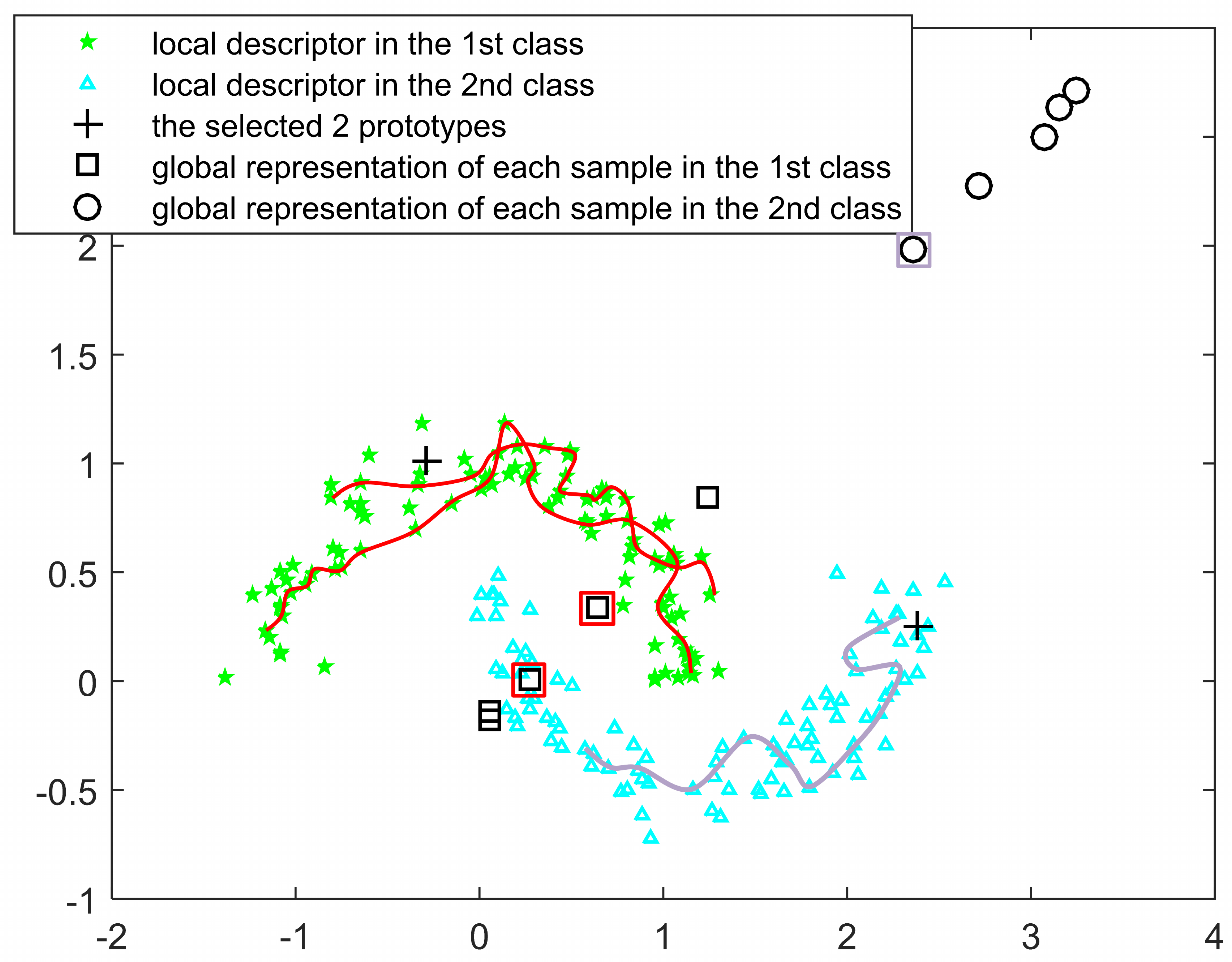}}
  \caption{Left: selection matrix $\bm Z$. Right: the aggregated linearly separable representations by ProLFA, where each curve represents 20 local descriptors included in each sample.}
  \label{fig:pro}
  \end{figure}

\begin{figure}[!t]
  \setlength{\abovecaptionskip}{0pt}
  \centering
  \graphicspath{{pics/}}
  {
  \label{fig:con}
  \includegraphics[width = 2.5in]{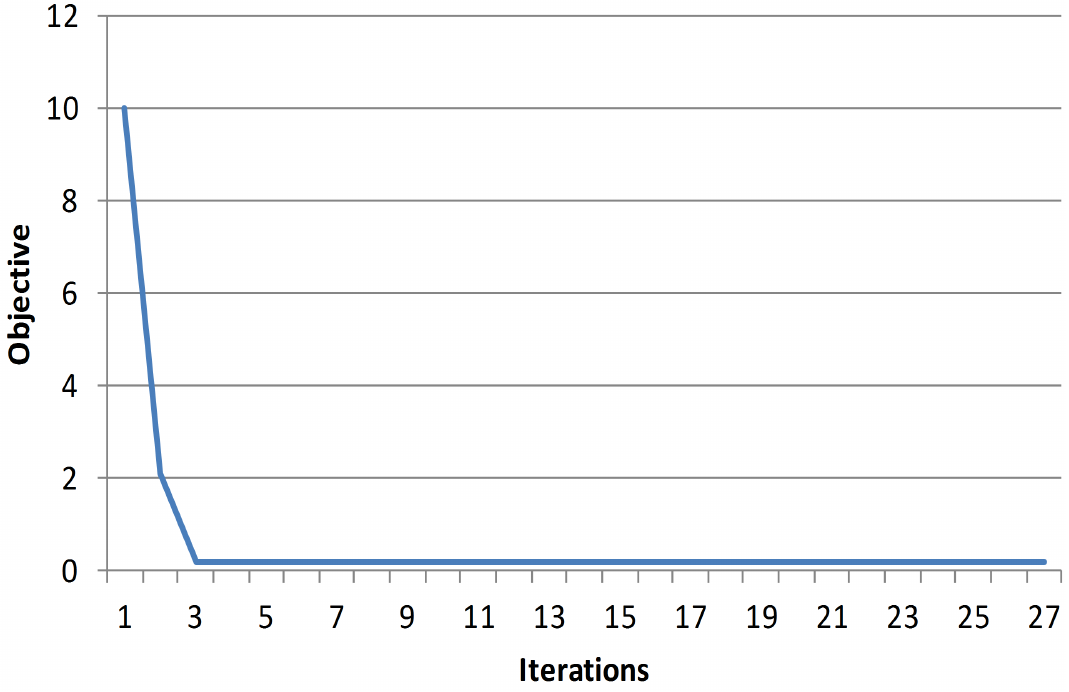}}
  \hspace{0.05cm}
 {
  \label{fig:com}
  \includegraphics[width = 2.5in]{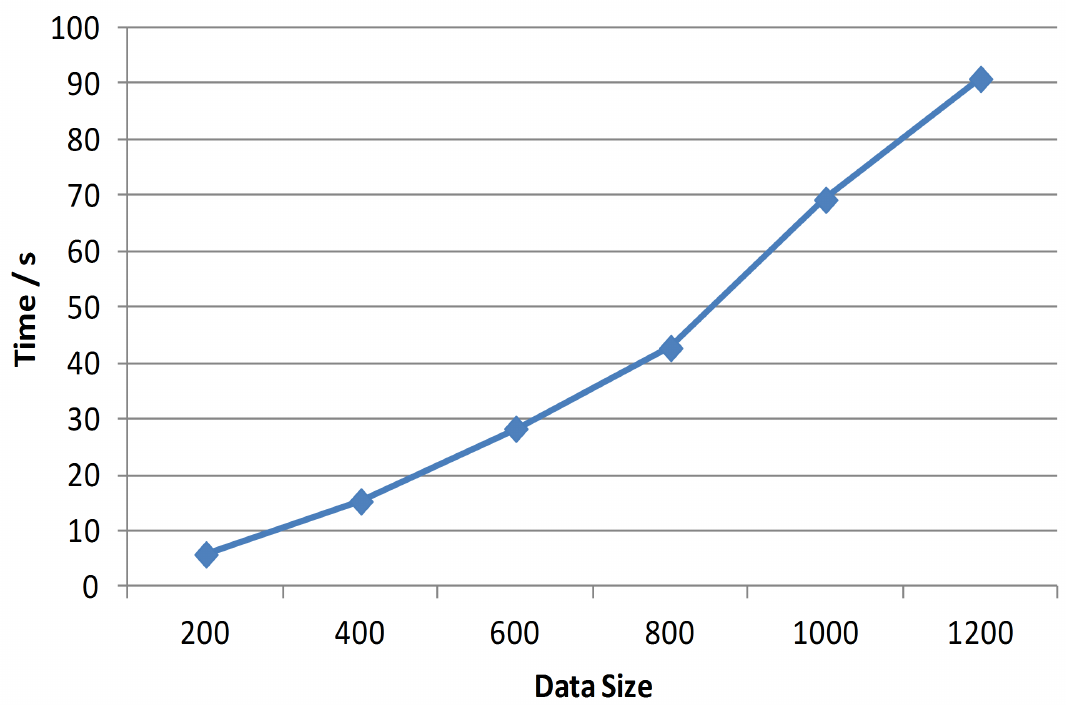}}
  \caption{Left: convergence speed. Right: training time of ProLFA.}
  \label{fig:per}
  \end{figure}

Figure~\ref{fig:pro} visualizes the selection matrix $\bm Z$ about 2 prototypes, and aggregated results by ProLFA on an artificial two-class dataset.
Specifically, the 200 points (\emph{i.e.,} local descriptors) in the dataset are first randomly grouped into 10 samples, and then we can obtain a global representation for each sample\footnote{The \textcolor{black}{source} code can be found at \url{https://github.com/indussky8/demo_ProLFA.}}.
It can be observed that the exclusivity property of selection matrix enhances the representativeness of prototypes, thus promoting the discrimination of aggregated representations. Although the convergence and complexity of ProLFA have been theoretically provided, it would be more intuitive to see its empirical behavior. Thus, we have shown the training speed and time of this synthetic dataset in Figure~\ref{fig:per}, where $\bar{d}=2$ and $P=1$.
Here, a roughly linearly increasing timing result is also consistent with that in Complexity Analysis.

\begin{table*}[!t]
\caption{Statistics for four datasets.} 
\label{tab:dataset}
\tiny
\begin{center}
\begin{tabular}{cccccccc}
    \hline \hline
     Dataset & \# images  & \#  classes  & Class imbalance? &  Size of images & \# SIFT/image   & \# DAISY/image  & $\bar{d}$ \\
     \hline
     Oxford5k & 5,062 & 11 & No & 1024 $\times$ 768 & 90 & 300 & 2048 \\
     INRIA Holidays  & 1,491 & 500 & No & 1024 $\times$ 768 &  90 & 300 & 2048 \\
     Fifteen Scene Categories & 4,485 & 15 & Yes &300 $\times$ 250 &   90 & 150 & 4096 \\
     Pascal VOC 2007  & 2,989 & 20 & Yes &300 $\times$ 300 &  90 & 150 & 4096 \\
    \hline\hline
    \end{tabular}
\end{center}
\end{table*}

\subsection{Evaluation by Image Search}

We assess the interpretability of aggregated representations by evaluating their search performance on \textbf{Oxford5k}~\cite{philbin2007object} and \textbf{INRIA Holidays}~\cite{jegou2010improving} datasets. 
\textcolor{black}{
The statistics about all the datasets is provided in Table~\ref{tab:dataset}.
we can find that such two datasets for search task are class balanced. Specially, the downloaded images are all rescaled to 1024 $\times$ 768, since the initial size is various, such as 1024 $\times$ 759 and 581 $\times$ 1024.
}
Specifically, Oxford5k consists of 5062 images of buildings and 55 query images corresponding to 11 distinct buildings in Oxford. 
The search quality is measured by the mean average precision (mAP) computed over the 55 queries.
\textcolor{black}{
mAP is also the widespread use in evaluating image search system~\cite{jegou2010improving} since it evaluates the average performance on all classes.
}
Holidays includes 1491 photos of different locations and objects, 500 of them being used as queries. The search quality is measured by mAP, with the query removed from the ranked list. 
Image search task is just used to evaluate the performance of the semi-supervised feature aggregation model in~(\ref{equ:ModelSemi}). Thus we annotated a small subset of Oxford5K (resp. Holidays), but not picking from another one. Such setup is also common in the scenario of semi-supervised image retrieval~\cite{wang2010semi}. 
Consequently, we randomly select $20\%$ of dataset in each class as the annotated samples. 
For each image in Oxford5K and Holidays, we take 90 and 300 descriptors as the input of each aggregation approach, respectively.
Then our Semi-ProLFA model in~(\ref{equ:ModelSemi}) is employed to train the aggregation function, thus obtaining the global representations of dataset via~(\ref{equ:reptest}). Table~\ref{tab:search} presents the compared search results produced in two kinds of descriptors, where the prototype size is 2048. 
As expected, our Semi-ProLFA model performs better compared with many representative unsupervised and fully supervised feature aggregation approaches. This is because in a semi-supervised aggregation way, representative prototypes can be selected with properties of diversity and discrimination. Thus, the aggregated representations are provided with more interpretability in search task. \textcolor{black}{Additionally, our Semi-ProLFA model achieves more promising results compared with ProLFA, since the proportion of annotated samples is very small. Thus, semi-supervised approaches can take more advantages in this case.}

 \begin{table}[!t]
 \footnotesize
 \begin{center}
 \caption{Impact of our method on search performance. The methods in \uppercase\expandafter{\romannumeral1}, \uppercase\expandafter{\romannumeral2} and \uppercase\expandafter{\romannumeral3} aim at aggregating local descriptors in unsupervised, fully supervised and semi-supervised scenarios, respectively.} \label{tab:search} 
 \begin{tabular}{cc|cc|cc}
 \hline
 \hline
   \multicolumn{2}{c|}{\multirow {3}{*}{Method $\downarrow $}} & \multicolumn{4}{c}{mAP}\\
   \cline{3-6}
   & &\multicolumn{2}{c|}{Oxford5k} &\multicolumn{2}{c}{Holidays}\\
   \cline{3-6}
   & &SIFT &DAISY 	&SIFT 	&DAISY\\
  \hline
  \multirow{7}{*}{\shortstack{\uppercase\expandafter{\romannumeral1}}} & BoW~\cite{csurka2004visual} & 51.35$\pm$0.34 &56.68$\pm$0.12 &	56.24$\pm$0.45 & 	56.22$\pm$0.16 	\\
  & VLAD~\cite{jegou2010aggregating} & 58.31$\pm$0.40 &59.33$\pm$0.18 &	56.34$\pm$0.62 & 	55.36$\pm$0.25 	\\
  & FV~\cite{perronnin2010improving} & 59.07$\pm$0.33 &56.31$\pm$0.19 &	60.34$\pm$0.25 & 	58.17$\pm$0.18 	\\
  & SC~\cite{ge2013sparse} & 63.69$\pm$0.55 &61.12$\pm$0.21 &	59.31$\pm$0.44 & 	61.36$\pm$0.44 	\\
  & $\phi_{\Delta }+\psi_{d} + \rm RN$~\cite{jegou2014triangulation} &
  	61.10$\pm$0.38 & 	67.14$\pm$0.37  & 73.39$\pm$0.62 &\textbf{76.12$\pm$0.40 }	\\
  & DM~\cite{furuya2015diffusion} & 59.66$\pm$ 0.18 & 55.64$\pm$0.14 &	59.69$\pm$0.32 & 	61.37$\pm$0.14 	\\
  \hline
  \multirow{6}{*}{\shortstack{\uppercase\expandafter{\romannumeral2}}}& $T_{1}\left ( \cdot  \right )$~\cite{katharopoulos2017learning} & 62.64$\pm$0.26 &60.33$\pm$0.37 &	66.01$\pm$0.18 & 	69.14$\pm$0.42 	\\
   & UniVCG~\cite{yang2008unifying} & 53.11$\pm$0.58 &51.03$\pm$0.56 & 54.86$\pm$0.29 & 58.32 $\pm$0.60 	\\
  &LC-KSVD2~\cite{jiang2013label}& 56.82$\pm$0.43 &56.11$\pm$0.36 & 62.81$\pm$0.10 & 60.54 $\pm$0.21 	\\
  & DBoWs~\cite{iosifidis2014discriminant} & 53.30$\pm$0.45 & 55.90 $\pm$0.50 & 59.31$\pm$0.58 & 	54.15$\pm$0.32 	\\
  & EO-BoW~\cite{passalis2016entropy} & 62.32$\pm$0.12 & 62.88$\pm$0.47 &	61.21$\pm$0.68 & 	62.53$\pm$0.52 	\\
  & ProLFA & 63.40$\pm$0.19 & 61.22$\pm$0.34 &	66.89$\pm$0.66 & 	66.94$\pm$0.41 	\\
  \hline
  \uppercase\expandafter{\romannumeral3}& Semi-ProLFA & \textbf{69.90$\pm$0.63} &\textbf{70.70$\pm$0.30} &	\textbf{75.33$\pm$0.28} & 	74.89$\pm$0.38 	\\
  \hline
  \hline
 \end{tabular}
 \end{center}
 \vspace{-0.3cm}
 \end{table}

\textcolor{black}{
In essence, $\phi_{\Delta }+\psi_{d} + \rm RN$~\cite{jegou2014triangulation} only obtains 1.23\% improvement over the strongest competitor (\emph{i.e.,} our Semi-ProLFA) on DAISY descriptor of Holidays dataset, while consistently performs worse on other cases. This is mainly due to the different data. Specifically, on such a small size dataset (Holidays), the advantage of semi-supervision is limited. In addition, we provide 300 DAISY descriptors for each photo, but 90 for another dataset.
Consequently, $\phi_{\Delta }+\psi_{d} + \rm RN$~\cite{jegou2014triangulation}, as an unsupervised method, can leverage more local information though no labels, thus performing best on this dataset even than ours.
}

 \begin{table}[!t]
 \footnotesize
 \begin{center}
 \caption{Impact of our method on classification task. The methods in \uppercase\expandafter{\romannumeral1}, \uppercase\expandafter{\romannumeral2} and \uppercase\expandafter{\romannumeral3} aim at aggregating local descriptors in unsupervised, fully supervised and semi-supervised scenarios, respectively.} \label{tab:class}
 \begin{tabular}{cc|cc|cc}
 \hline
 \hline
   \multicolumn{2}{c|}{\multirow {3}{*}{Method $\downarrow $}} & \multicolumn{4}{c}{Accuracy}\\
   \cline{3-6}
   & &\multicolumn{2}{c|}{Fifteen Scene} &\multicolumn{2}{c}{Pascal VOC 2007}\\
   \cline{3-6}
    & &SIFT &DAISY 	&SIFT 	&DAISY\\
  \hline
  \multirow{7}{*}{\shortstack{\uppercase\expandafter{\romannumeral1}}} & BoW~\cite{csurka2004visual} & 66.99 $\pm$0.56 &64.88$\pm$0.52 &	43.90$\pm$0.44 & 	41.93$\pm$0.71 	\\
  & VLAD~\cite{jegou2010aggregating} & 71.93 $\pm$0.50 &69.41$\pm$0.36 &	46.69$\pm$0.21 & 	49.38$\pm$0.14 	\\
  & FV~\cite{perronnin2010improving} & 70.12 $\pm$0.21 &70.73$\pm$0.17 &	48.36$\pm$0.14 & 	50.03$\pm$0.31 	\\
  & SC~\cite{ge2013sparse} & 72.20 $\pm$0.15 &74.69$\pm$0.21 &	49.08$\pm$0.14 & 	48.58$\pm$0.50	\\
  & $\phi_{\Delta }+\psi_{d} + \rm RN$~\cite{jegou2014triangulation} & 77.41$\pm$0.43 &76.30$\pm$0.78 &	49.69$\pm$0.68 & 	50.29$\pm$0.36 	\\
  & DM~\cite{furuya2015diffusion} & 69.36$\pm$0.57 &71.21$\pm$0.56 &	41.63$\pm$0.23 & 	40.30$\pm$0.40 	\\
  \hline
  \multirow{6}{*}{\shortstack{\uppercase\expandafter{\romannumeral2}}} 
  & $T_{1}\left ( \cdot  \right )$~\cite{katharopoulos2017learning} & 73.74$\pm$0.21 &74.86$\pm$0.11 &	46.26$\pm$0.43 & 	48.18$\pm$0.59 	\\
  & UniVCG~\cite{yang2008unifying} & 72.94$\pm$0.20 &68.07$\pm$0.50 &	47.41$\pm$0.21 & 	49.55$\pm$0.16 	\\
  & LC-KSVD2~\cite{jiang2013label}& 78.35 $\pm$0.19 &75.17$\pm$0.42 & 47.83 $\pm$0.33 & 54.52 $\pm$0.36 	\\
  & DBoWs~\cite{iosifidis2014discriminant} & 73.49$\pm$0.17 & 72.10$\pm$0.24 &	47.29$\pm$0.59 & 	53.66$\pm$0.22 	\\
  & EO-BoW~\cite{passalis2016entropy} & 78.43$\pm$0.43 & 75.60$\pm$0.42 &	49.30$\pm$0.14 & 	55.25$\pm$0.17 	\\
  & ProLFA & \textbf{80.88 $\pm$0.39} &\textbf{80.33$\pm$0.21} &	\textbf{52.50$\pm$0.48}& 	\textbf{56.94$\pm$0.27} 	\\
  \hline
  \uppercase\expandafter{\romannumeral3}& Semi-ProLFA & 76.10 $\pm$ 0.19 & 77.81 $\pm$0.32 & 49.76$\pm$0.42 & 	54.01$\pm$0.30 	\\
  \hline
  \hline
 \end{tabular}
 \end{center}
 \vspace{-0.3cm}
 \end{table}
  
\subsection{Evaluation by Image Classification}~\label{EvaluateByClass}

We assess the discrimination of aggregated representations by evaluating their classification performance on \textbf{Fifteen Scene Categories dataset}~\cite{Lazebnik2006Beyond} and \textbf{Pascal VOC 2007}~\cite{li2013adaptive}. 
\textcolor{black}{
As shown in Table~\ref{tab:dataset}, the two datasets used for classification task are slightly imbalanced. Concretely, for Fifteen Scene Categories dataset, each category has 200 to 400 images, and average image size is 300 $\times$ 250 pixels. While for Pascal VOC 2007, each category has 96 to 600 images with the general size of 500 $\times$ 375 or 375$\times$500, and we rescale all images to 300 $\times$ 300. 
}
Specifically, Scene dataset consists of both natural and man-made scenes, and has 4485 images in total. Pascal VOC 2007 is a widely used dataset for image classification. Here, a subset of its training and validation data that contains 2989 images from 20 object categories with one label is used as in~\cite{li2013adaptive}. Then, the evaluation protocol in~\cite{elhamifar2017subset} is used: $80\%$ of dataset is sampled from each class to build the training set, and the rest is used for testing.
For each image in Scene and Pascal, we take 90 and 150 descriptors as the input of each aggregation approach, respectively.
Our ProLFA model in~(\ref{equ:ModelRelax2}) is then employed to train the aggregation function on the training set, thus obtaining the global representations of testing set via~(\ref{equ:TestSample}). Finally, to evaluate the discrimination of aggregated representations, \textcolor{black}{we choose the 1-Nearest Neighbor (1-NN) classifier with Euclidean distance} since it is parameter free and the results will be easily reproducible. 
\textcolor{black}{
Naturally, the classification accuracy by 1-NN classifier is also used to evaluate all compared methods for a fair comparison.
}

Table~\ref{tab:class} presents the compared classification accuracy results produced in two kinds of descriptors, where the prototype size is 4096. As can be seen, our ProLFA outperforms many state-of-the-art unsupervised and fully supervised feature aggregation approaches. 
Additionally, our extension Semi-ProLFA can achieve comparable performance with that in fully supervised scenario. This is due to the fact that the proposed method improves the discrimination of aggregated representation by task-oriented prototype selection and domain-invariant projection.
\textcolor{black}{
However, for this task, ProLFA still consistently outperforms Semi-ProLFA, where the improvements range from 2.52\% to 4.78\%. This is just because of the larger proportion (80\%) of labeled data for each dataset. In this scenario, ProLFA, as a fully supervised method, takes more advantage than a semi-supervised method (Semi-ProLFA).
}

\begin{figure}[!t]
  \setlength{\abovecaptionskip}{0pt}
  \centering
  \graphicspath{{pics/}}
  \subfigure[]
  {
  \label{fig:scene_SIFT}
  \includegraphics[width = 2.50in]{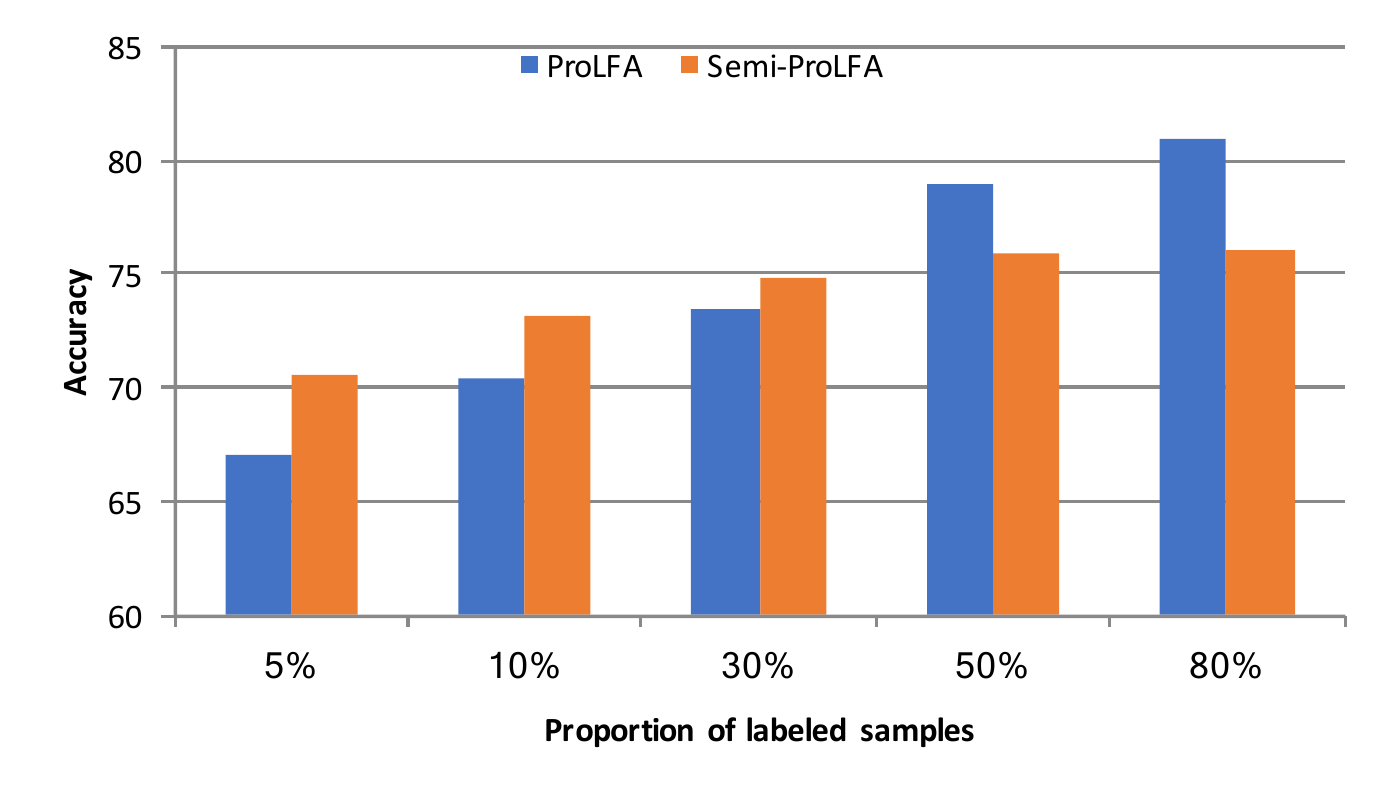}}
  \hspace{0.03cm}
  \subfigure[]
  {
  \label{fig:scene_DAISY}
  \includegraphics[width = 2.50in]{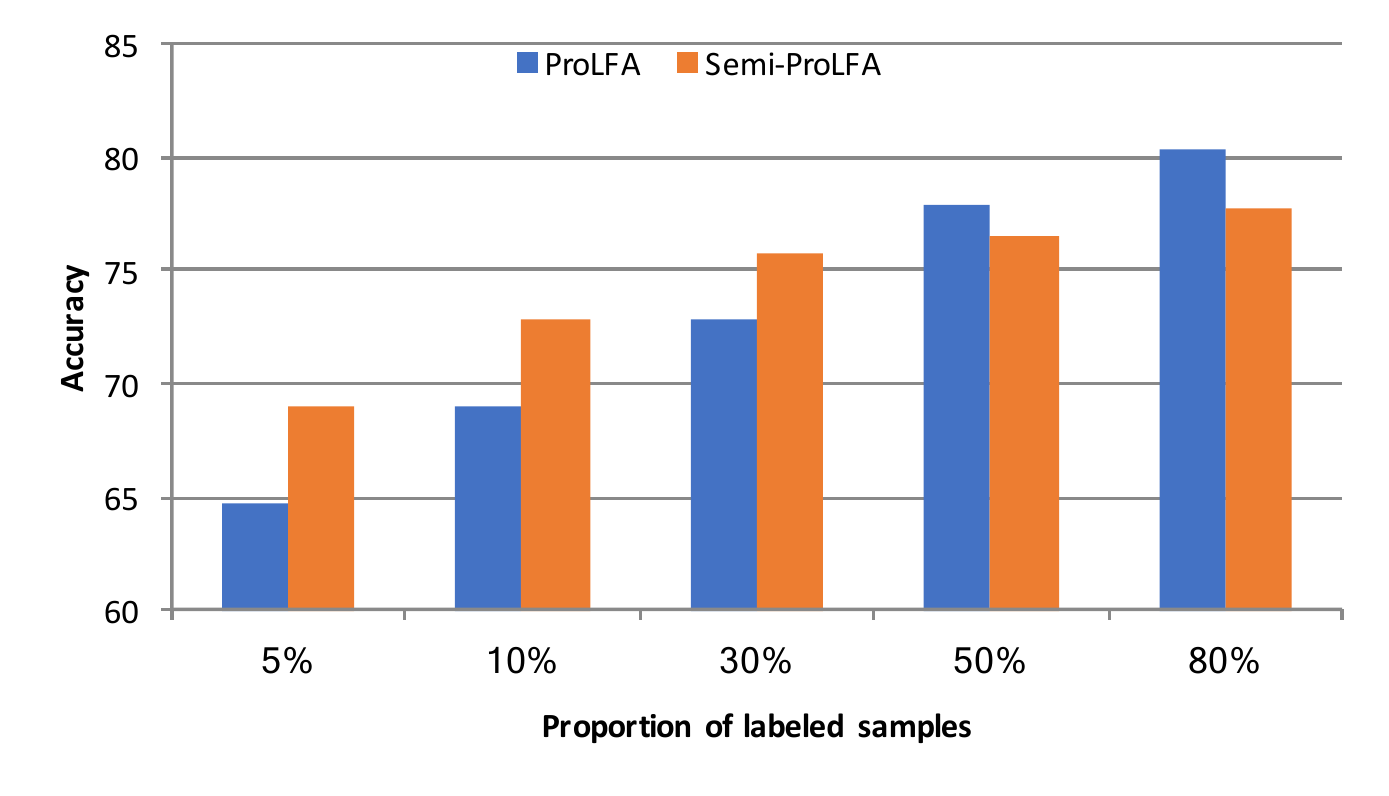}}
  \caption{Impact of labeled proportion in Scene dataset by Semi-ProLFA and ProLFA. (a) on SIFT. (b) on DAISY.}
  \label{fig:scene_propo}
  \end{figure}

Next, we compare the performances of our ProLFA and Semi-ProLFA with respect to different proportions of labeled samples in Fifteen Scene Categories dataset. As shown in Figure~\ref{fig:scene_propo}, Semi-ProLFA performs better in the case of a few labeled samples, which means that it is suitable to perform semi-supervised local feature aggregation. While on sufficient labeled samples, ProLFA can perform better. 
\textcolor{black}{
In addition, as summarized in Table~\ref{tab:dataset}, the four datasets have covered all the possibilities, including class balance, slight and severe class imbalance. However, as reported in Table~\ref{tab:search} and Table~\ref{tab:class}, our method (Semi-ProLFA and ProLFA) can outperform other competitors in most cases. This just show that the class imbalance may have no obvious impact on the advantage of our method.
}

\textcolor{black}{
To present the efficiency of each method, we additionally report physical running time to generate global representations of Fifteen Scene Categories dataset, where SIFT and DAISY descriptors serve as input of each method. All the times we report are estimated using a single CPU of a 2.3 GHz Xeon machine with 32 GB of RAM. 
As shown in Table~\ref{tab:timeOfmethod}, we can find that the complexity of our method is a general case in feature aggregation, although is not the lowest. Besides, supervised methods task more time than unsupervised ones generally, since they need to additionally train a model on labeled data. 
In particular, the added overhead can be neglected in practice to the significant accuracy improvements achieved by our ProLFA. }

\begin{table}[t!]
    \centering
    \tiny
    \caption{Physical running time (\emph{sec.}) to generate global representations of Fifteen Scene Categories dataset by various methods.}
    \label{tab:timeOfmethod}
    \begin{tabular}{c|cccccc|c}
    \hline
    \hline
    \multirow{2}{*}{\textbf{Method}}  & \multicolumn{6}{c|}{\textbf{Unsupervised}} &     \\
    \cline{2-8}
    &BoW~\cite{csurka2004visual}& VLAD~\cite{jegou2010aggregating}& FV~\cite{perronnin2010improving}& SC~\cite{ge2013sparse}& $\phi_{\Delta }+\psi_{d} + \rm RN$~\cite{jegou2014triangulation}& DM~\cite{furuya2015diffusion}\\
    \hline %
    \textbf{SIFT}
    & 208.9 & 314.4 & 384.1 & 293.1 & 314.5 & 301.3  &  \\
    \textbf{DAISY}
    & 194.8 & 319.2 & 358.2 & 285.7 & 322.9 & 333.9  &  \\
    \hline
     \multirow{2}{*}{\textbf{Method}}  &  \multicolumn{6}{c|}{\textbf{Supervised}}  & \textbf{Semi-supervised}     \\
    \cline{2-8}
    &$T_{1}\left ( \cdot  \right )$~\cite{katharopoulos2017learning}& UniVCG~\cite{yang2008unifying}& LC-KSVD2~\cite{jiang2013label}& DBoWs~\cite{iosifidis2014discriminant}& EO-BoW~\cite{passalis2016entropy}& ProLFA  & Semi-ProLFA \\
    \hline %
    \textbf{SIFT}
    & 834.1 & 1042.2 & 688.2 & 1233.4 & 941.7 & 1023.3 & 1001.1  \\
    \textbf{DAISY}
    & 794.1 & 1032.2 & 633.5 & 1183.4 & 978.6 & 984.5 & 1030.3  \\
    \hline
    \hline
    \end{tabular}
\end{table}

  \begin{figure}[!t]
  \setlength{\abovecaptionskip}{0pt}
  \centering
  \graphicspath{{pics/}}
  \subfigure[]
 {
  \label{fig:D1}
  \includegraphics[width = 2.50in]{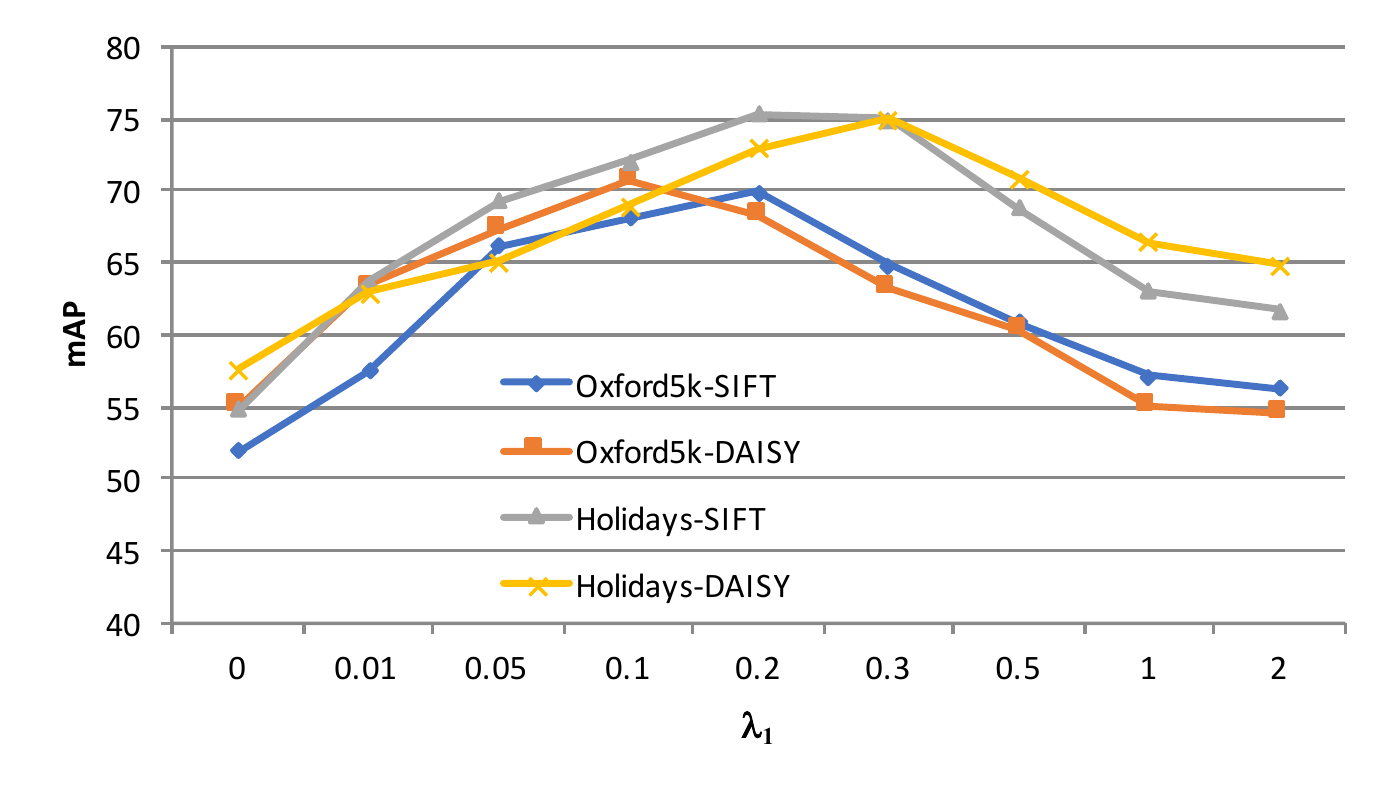}}
  \hspace{0.03cm}
  \subfigure[]
 {
  \label{fig:D3}
  \includegraphics[width = 2.50in]{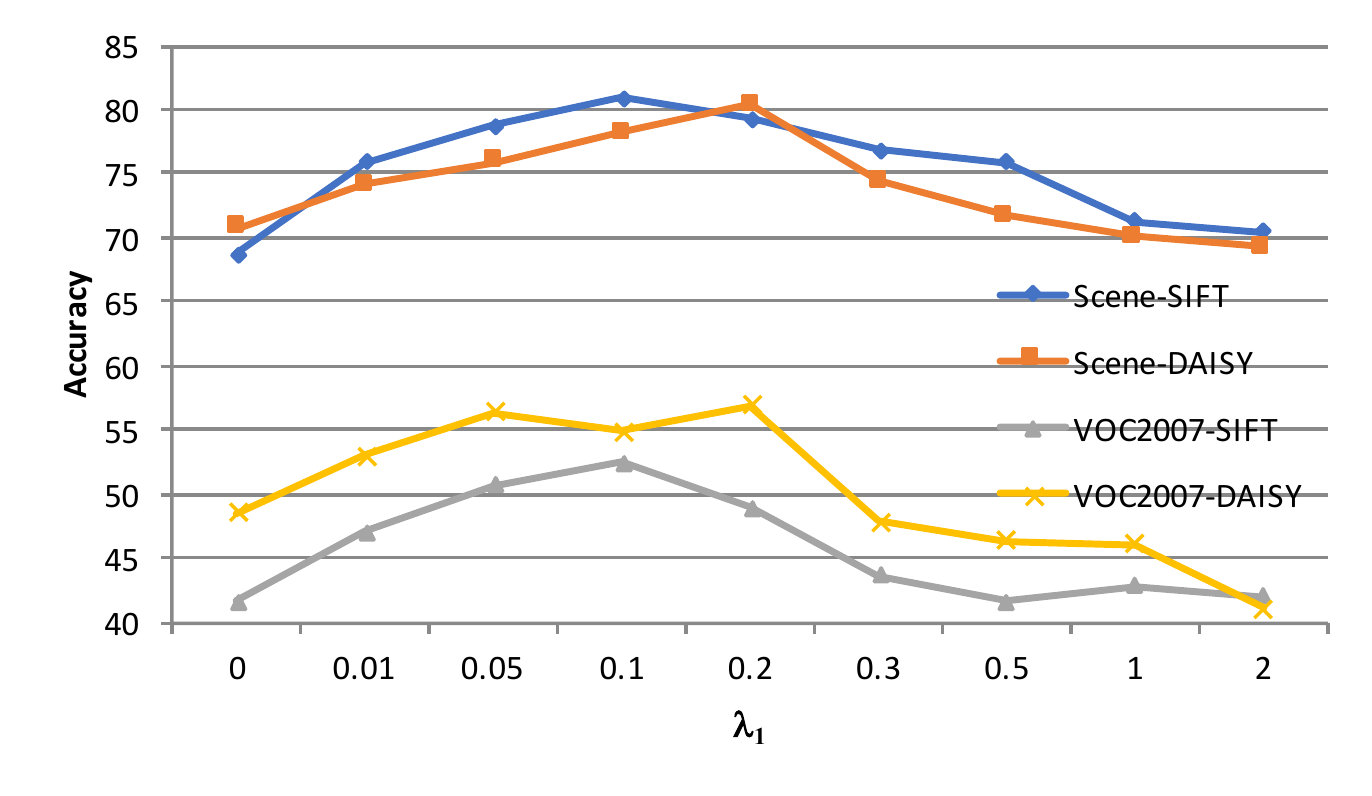}}
  \caption{Impact of $\lambda_{1}$ on the two tasks. (a) Image search by Semi-ProLFA. (b) Image classification by ProLFA.}
  \label{fig:lambda}
  \end{figure}

  \begin{figure}[!t]
  \setlength{\abovecaptionskip}{0pt}
  \centering
  \graphicspath{{pics/}}
  \subfigure[]
  {
  \label{fig:search}
  \includegraphics[width = 2.50in]{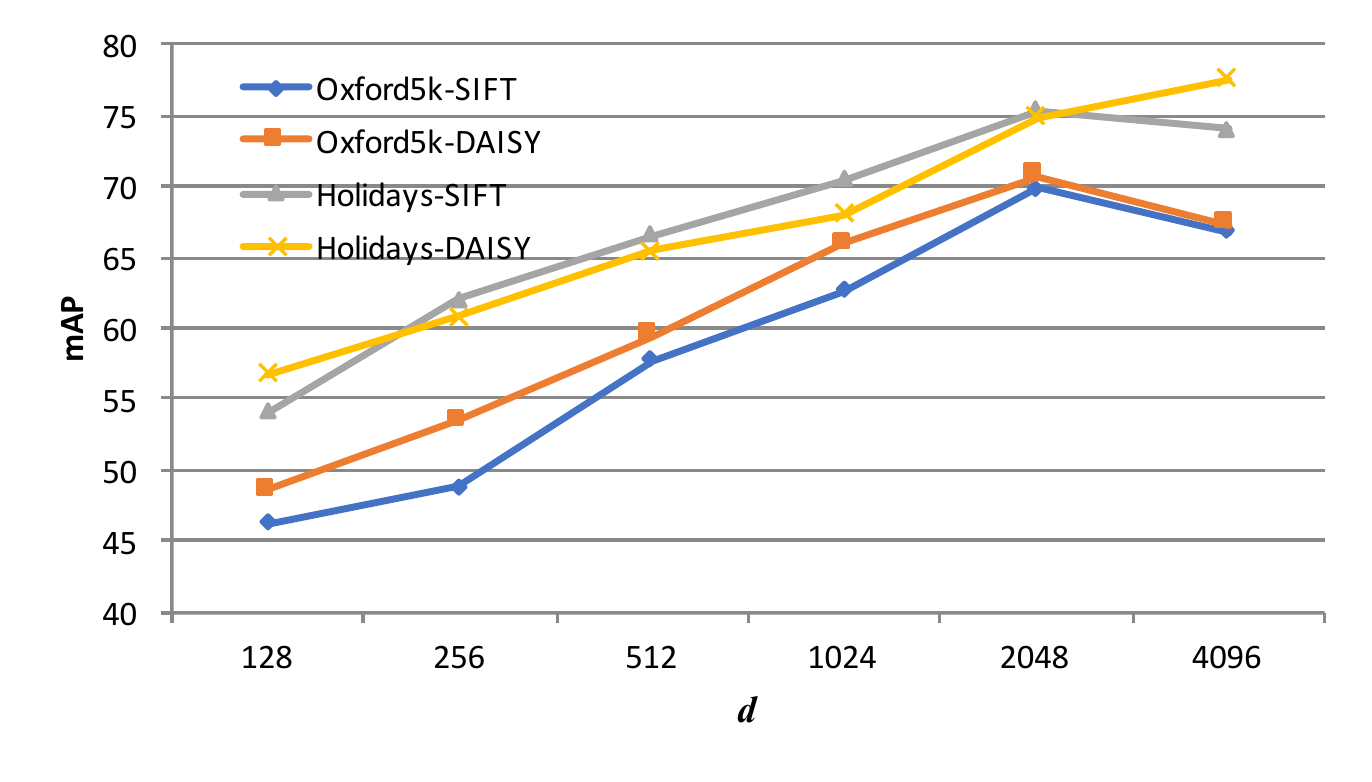}}
  \hspace{0.03cm}
  \subfigure[]
  {
  \label{fig:class}
  \includegraphics[width = 2.50in]{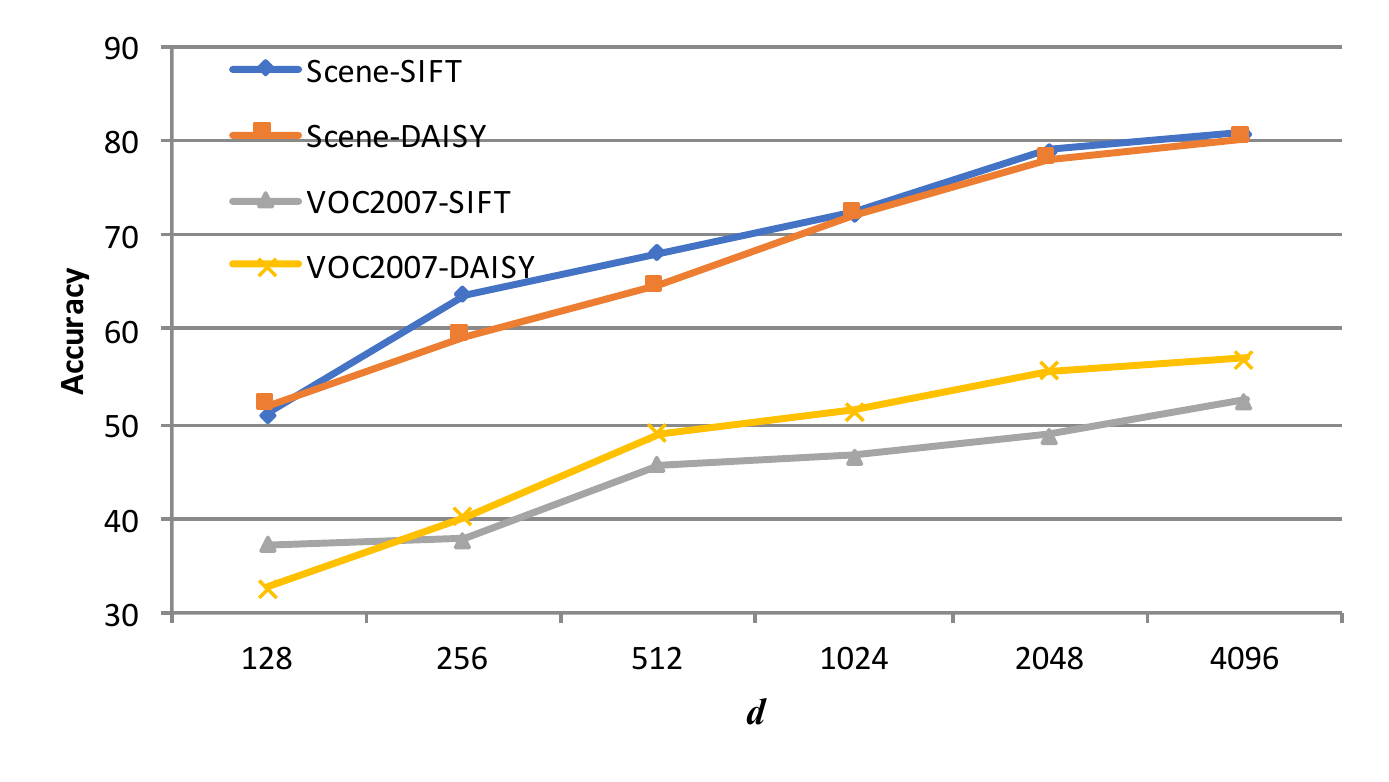}}
  \caption{Impact of $\bar{d}$ on the two tasks. (a) Image search by Semi-ProLFA. (b) Image classification by ProLFA.}
  \label{fig:d}
  \vspace{-0.45cm}
  \end{figure}
  
\subsection{Parameter Analysis}

The main parameters in our ProLFA model are prototype size $\bar{d}$, and the regularization parameter $\lambda_{1}$ ($\lambda_{2}$ is relatively less important). The analysis of these parameters is shown in Figure~\ref{fig:lambda} and Figure~\ref{fig:d} for search and classification tasks. The conclusions drawn are identical on both tasks. For all datasets, the performance is an increasing function of the prototype size, but there indeed exists a turning point, which is around $N/6.5$ in scene categorization task. The general trend for other tasks needs to be further studied.
In addition, as $\lambda_{1}$ grows, the mAP drops after growing. This is because firstly the prototypes are selected more discriminatively, and secondly less and less effort is put on fitting data.

\textcolor{black}{
It is worth noting that there only exist 6 times of execution for each method, and thus a Wilcoxon test is also necessary to claim our competitiveness against the methods compared.
Concretely, at the default 5\% significance level, we conduct a paired, two-sided Wilcoxon signed rank test on the six pairs of results about our approach and each compared aggregation approach using the exact method. For search task, we perform Wilcoxon test on our Semi-ProLFA and each competitor, while on ProLFA and each competitor for classification task.
The statistical results in terms of $p$-value of the test on the results of two tasks are almost {$0.0312$}, less than 0.05. Thus, there is enough statistical evidence to conclude that our method is indeed more competitive against all compared methods.
}

\begin{figure*}[t!]
\centering
\subfigure[On SIFT descriptors]{
\includegraphics[width=2.2in]{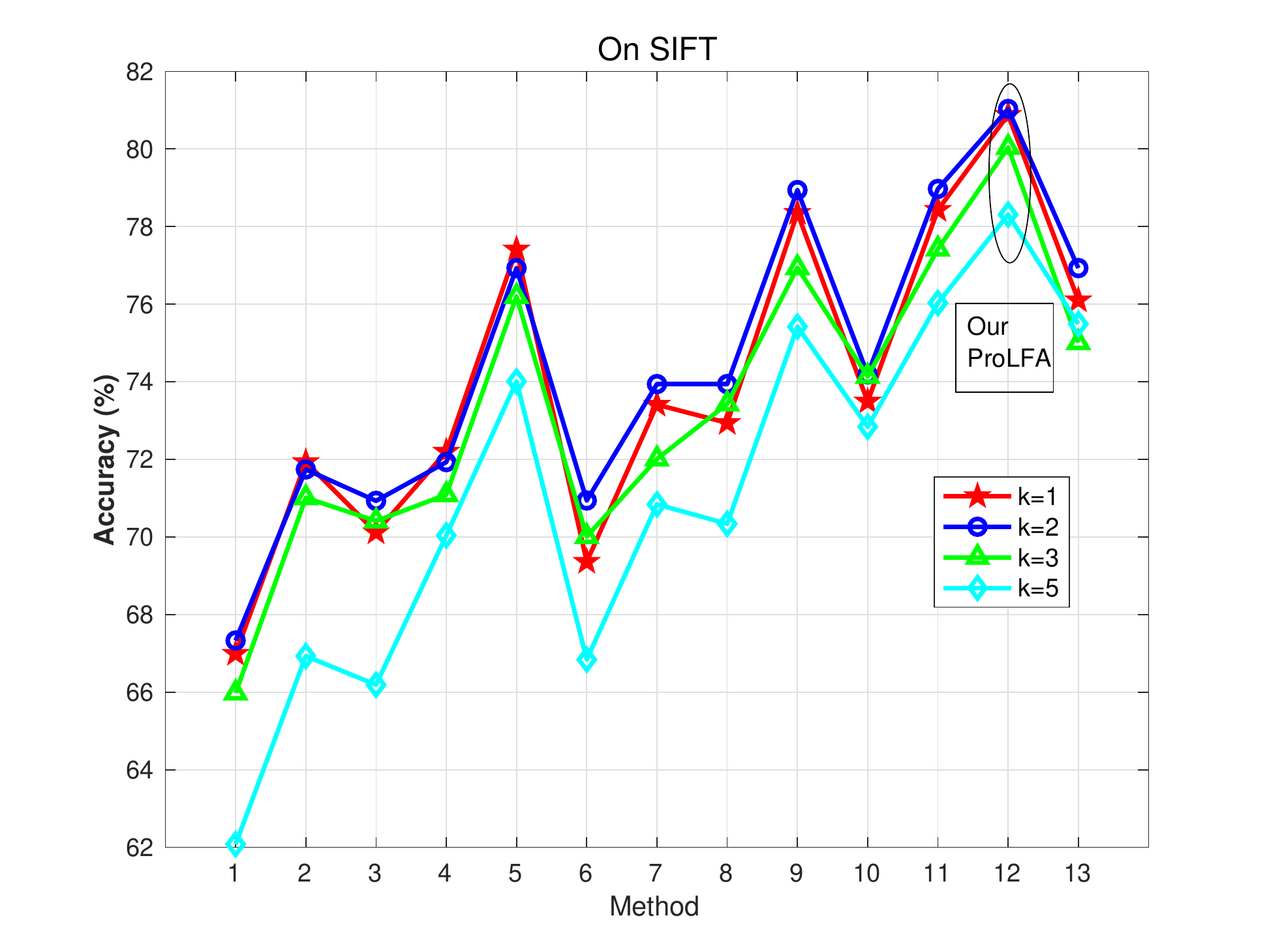}
\label{fig:k_sift}
}
\quad
\subfigure[On DAISY descriptors]{\label{fig:k_daisy}
\includegraphics[width=2.2in]{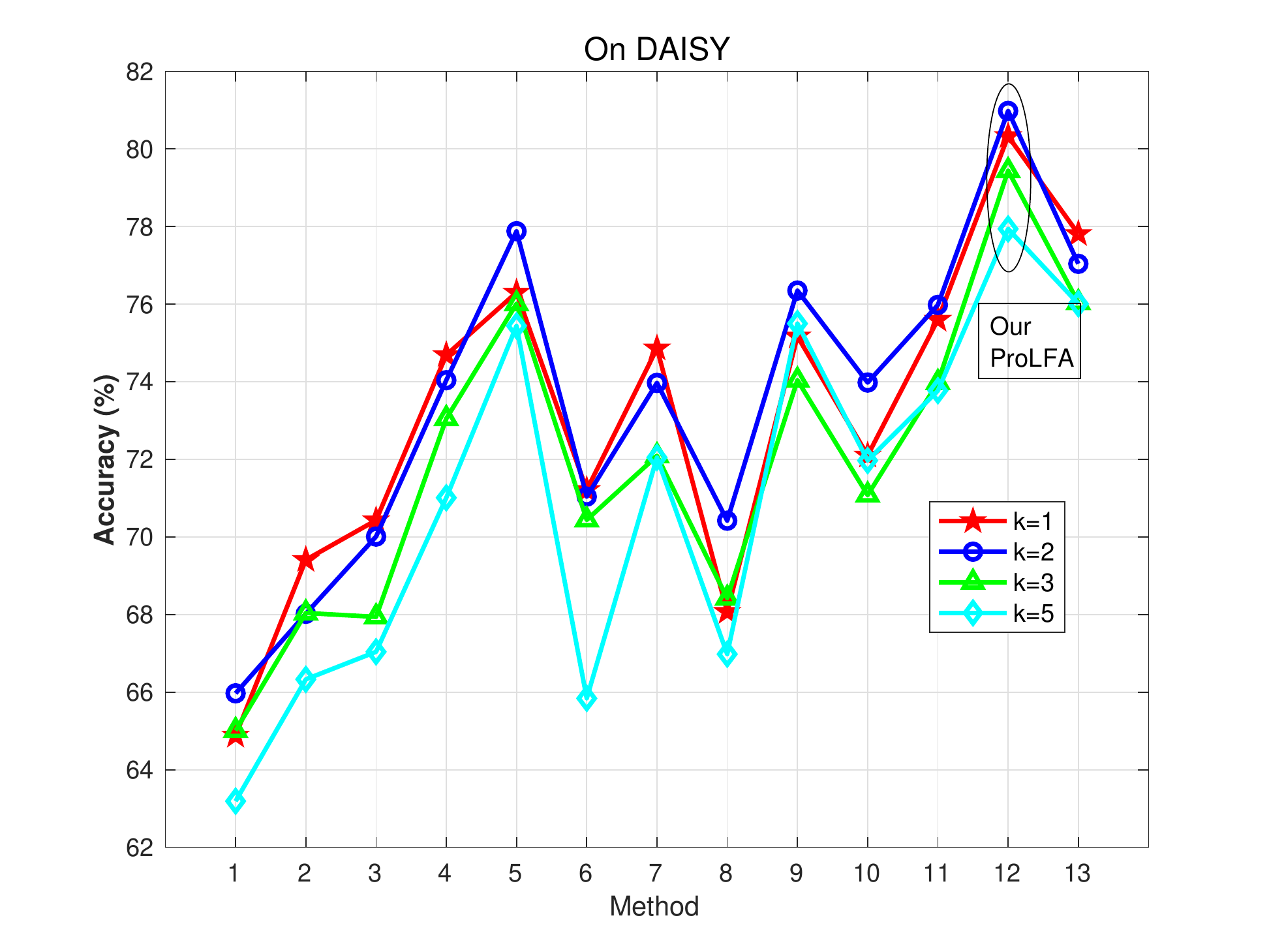}
}
\caption{Impact of `k' (neighbours) values on classification with Fifteen Scene dataset, where the numbers 1-13 in horizontal axis denote BoW~\cite{csurka2004visual}, VLAD~\cite{jegou2010aggregating}, FV~\cite{perronnin2010improving}, SC~\cite{ge2013sparse}, $\phi_{\Delta }+\psi_{d} + \rm RN$~\cite{jegou2014triangulation}, DM~\cite{furuya2015diffusion}, $T_{1}\left ( \cdot  \right )$~\cite{katharopoulos2017learning}, UniVCG~\cite{yang2008unifying}, LC-KSVD2~\cite{jiang2013label}, DBoWs~\cite{iosifidis2014discriminant}, EO-BoW~\cite{passalis2016entropy}, ProLFA, and Semi-ProLFA methods, respectively.}
\label{fig:kEffect}
\vspace{-0.5cm}
\end{figure*}

\begin{figure*}[t!]
\centering
\subfigure[On SIFT descriptors]{\label{fig:dis_sift}
\includegraphics[width=2.2in]{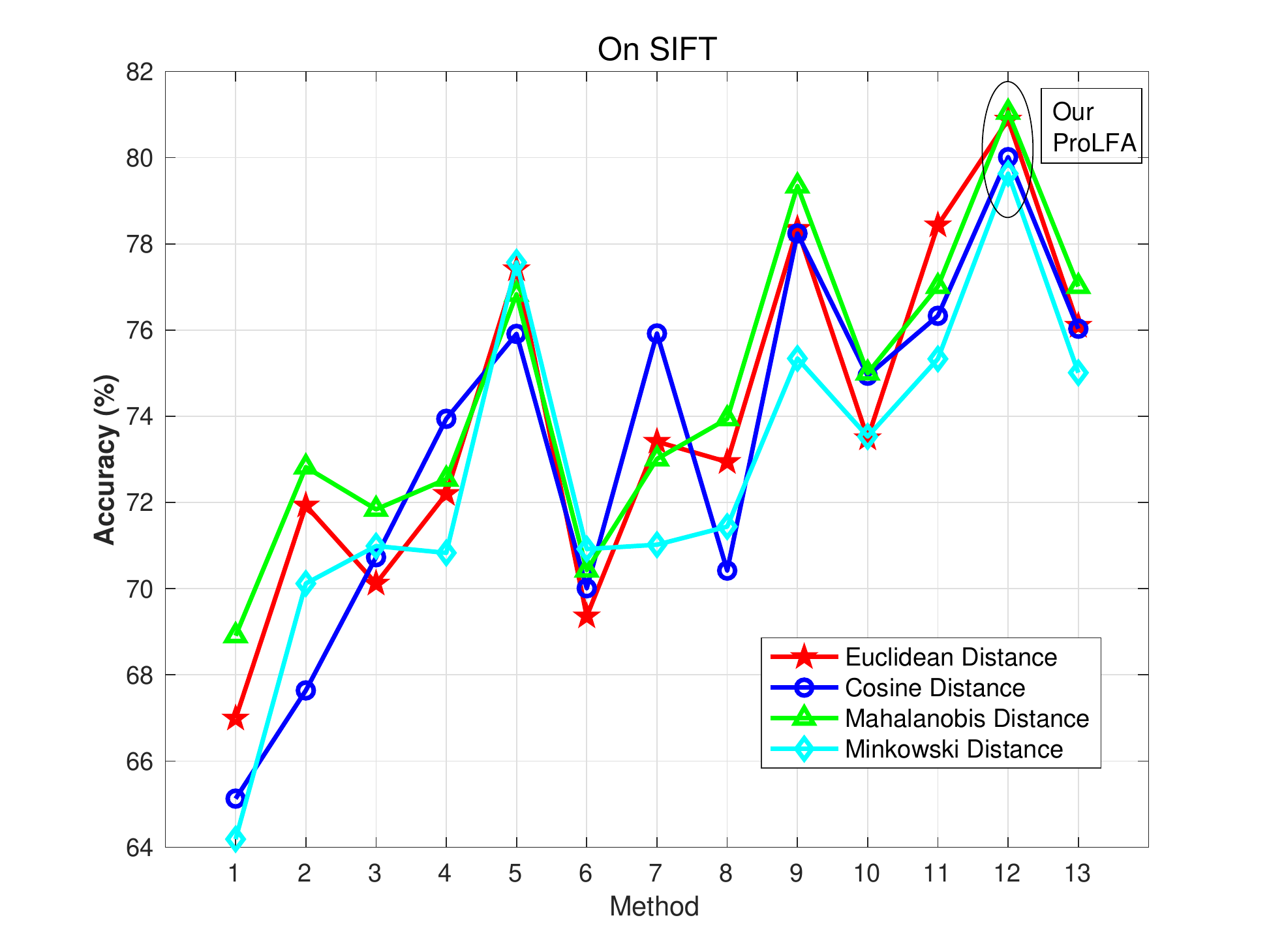}
}
\quad
\subfigure[On DAISY descriptors]{\label{fig:dis_daisy}
\includegraphics[width=2.2in]{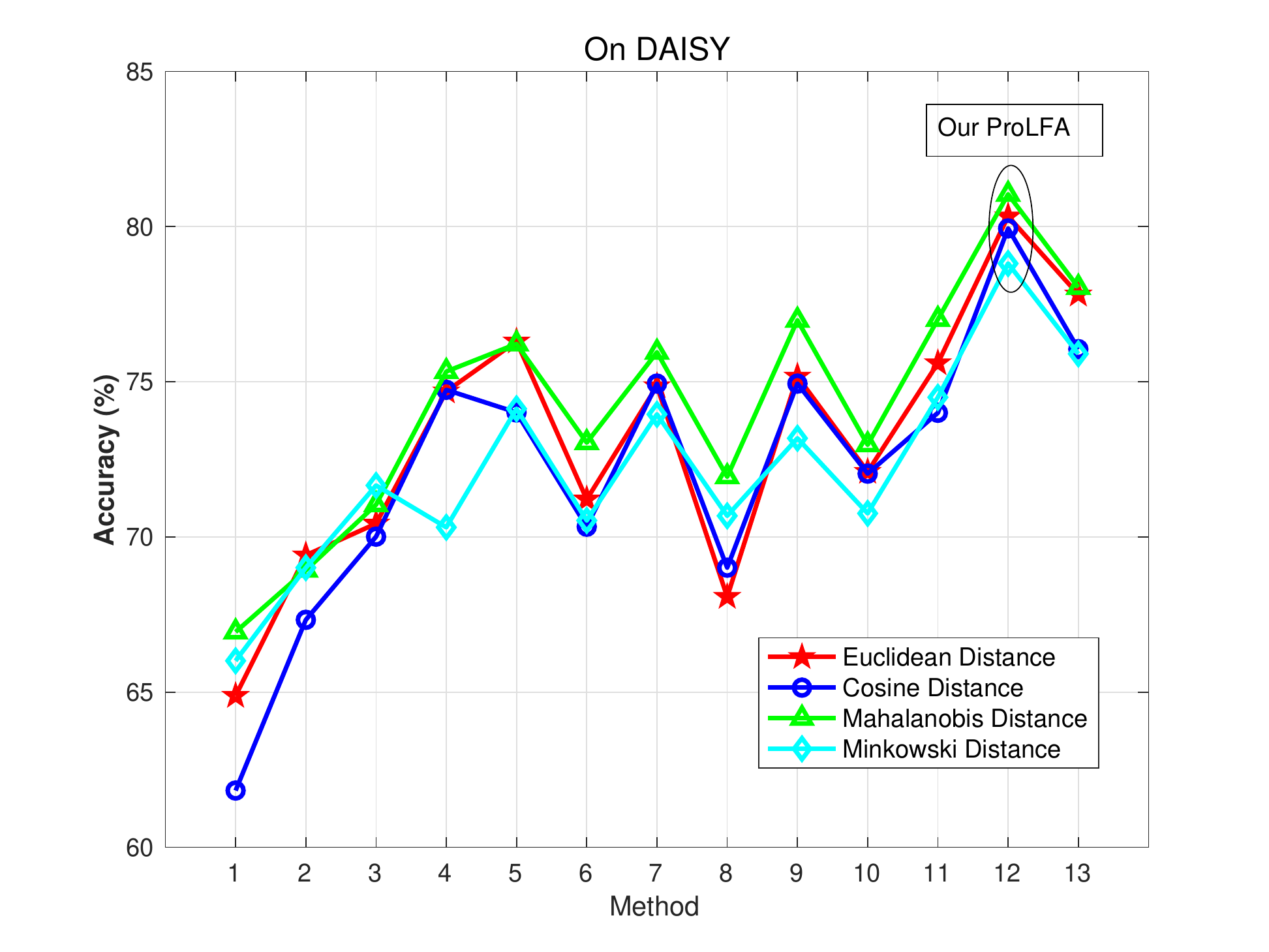}
}
\centering
\caption{Impact of four distance metrics on classification with Fifteen Scene dataset.}
\label{fig:DistanceEffect}
\end{figure*}

\textcolor{black}{
Furthermore, we conduct two additional studies about NN classifier on Fifteen Scene Categories dataset: (i) the effect of using different `k' values (including $k=\{1,2,3,5\}$) by using Euclidean distance in NN classifier; (ii) the effect of using different distances (including Euclidean, Cosine, Mahalanobis, and Minkowski Distances) in 1-NN classifier.
The results are reported in Figure~\ref{fig:kEffect} and Figure~\ref{fig:DistanceEffect}, respectively.
As expected, our ProLFA can achieve the best classification accuracy with different neighbours values and distance metrics. However, `k' and distance metric have a slight effect on our ProLFA. For example, it can be found from Figure~\ref{fig:kEffect} and Figure~\ref{fig:DistanceEffect} that $k>3$ and Minkowski distance would degrade the classification performance. But it is worth noting that the superiority of aggregated representation by our ProLFA is not influenced by these setups.
}

\textcolor{black}{
\subsection{Scalability Analysis}
To show the scalability of the proposed ProLFA to large-scale problems, we further conduct image classification on a larger dataset (Caltech-UCSD Birds)~\cite{welinder2010caltech}, which contains 11,788 images from 200 bird species.
Each species has approximately 30 images for training and 30 for testing. 
Totally, there are 5994 training images and 5794 test images.
In light of the good performance of CNN, we aggregate the second-order features using fine-tuned VGG-16~\cite{simonyan2014very} network. That is, we resize input images to $224 \times 224$ and aggregate the last convolutional layer features after ReLU activation. Thus, the size of local descriptors for each image is $28 \times 28 \times 512$ (\emph{i.e.,} $N_{i}=784$ for all $i$ and $d=512$).
We first test the performance of our ProLFA as in Section~\ref{EvaluateByClass}, where the prototype size is $4096$. 
Then, we compare our method with a recent approach named \textbf{$\gamma$-democratic}~\cite{lin2018second} and a neural network-based approach named \textbf{FV+NN}~\cite{perronnin2015fisher}.
For {$\gamma$-democratic}~\cite{lin2018second}, we directly adopt the aggregated features from their source code~\footnote{http://vis-www.cs.umass.edu/o2dp} with $\gamma=0.5$, while {FV+NN}~\cite{perronnin2015fisher} is reproduced by following FV encoding layer with a Multi-Layer Perceptron without data augmentation but with bagging. 
Finally, 1-NN classifier is used to evaluate the discrimination of aggregated features by each method.
Table~\ref{tab:largedata} presents the compared results. It can be observed that our ProLFA achieves a 3.3 \% improvement over FV+NN~\cite{perronnin2015fisher}, while a 1.8 \% deterioration over
$\gamma$-democratic~\cite{lin2018second}. 
This is mainly because unlike $\gamma$-democratic~\cite{lin2018second} that performs end-to-end fine-tuning, both FV+NN~\cite{perronnin2015fisher} and our ProLFA directly adopt the local CNN descriptors extracted in advance.
Extraction of local descriptors is also not the focus of this work.
However, due to task-specific prototype selection in feature aggregation, our ProLFA takes more advantages over FV+NN [C] that still adopts conventional FV aggregation.
}
 \begin{table}[!t]
 \footnotesize
 \begin{center}
 \caption{Comparison with two aggregation approaches on Caltech-UCSD Birds.} \label{tab:largedata}
 \begin{tabular}{c|ccc}
 \hline
 \hline
   Method & FV+NN~\cite{perronnin2015fisher} & $\gamma$-democratic~\cite{lin2018second} & ProLFA \\
   \hline
   Accuracy & 77.2 & 82.3 & 80.5\\
  \hline
  \hline
 \end{tabular}
 \end{center}
 \vspace{-0.5cm}
 \end{table}

\section{Conclusion and Future Work}

Local feature aggregation is a fundamental problem for numerous applications. In this work, we have introduced, first, a domain-invariant prototype selection based feature aggregation approach (ProLFA) to produce compact representations with the properties of interpretability and discrimination, and second, a composite Block Coordinate Descend (cBCD) framework to efficiently solve the proposed optimization program. Third, by experiments on different local features and tasks, we showed that ProLFA improves the state of the art on the problem of local feature aggregation, even in semi-supervised scenario. Finally, in light of the good performance of parallel optimization algorithms, ProLFA is provided with a potential scaling ability to very large datasets, which is also included in our ongoing research work.

\section*{Acknowledgments}
This work was supported in part by the National Key Research and Development of China under Grant 2016YFB0800404, in part by the National Natural Science Foundation of China under Grants 61532005, 61332012, and 61572068, and in part by the Fundamental Research Funds for the Central Universities under Grant 2018JBZ001.



 \newpage
 \section*{References}
 \bibliographystyle{elsarticle-num} 
 \bibliography{reference}





\end{document}